\documentclass[11pt]{article}

\usepackage[final]{acl}
\usepackage{times}
\usepackage{latexsym}
\usepackage[T1]{fontenc}
\usepackage[utf8]{inputenc}
\usepackage{microtype}
\usepackage{inconsolata}
\usepackage{graphicx}
\usepackage{amsmath}
\usepackage{amssymb}
\usepackage{booktabs}
\usepackage{multirow}
\usepackage{xcolor}
\usepackage{subcaption}
\usepackage{algorithm}
\usepackage{algorithmic}
\usepackage{subcaption}
\usepackage{listings}
\usepackage{tcolorbox}
\tcbuselibrary{breakable}
\usepackage[table]{xcolor}

\title{Compiler-Guided Adaptive Proof Search with Cross-Model Synergy on Context-Dependent Theorem Proving}

\author{Zhuo Liu \\
  University of Rochester \\
  \texttt{zhuo.liu@rochester.edu } \\\And
  Ding Yu \\
  University of Rochester \\
  \texttt{dyu18@ur.rochester.edu} \\\And
  Hangfeng He \\
  University of Rochester \\
  \texttt{hangfeng.he@rochester.edu} \\}

\begin{document}
\maketitle

\begin{abstract}
Theorem proving in real-world Lean~4 projects is challenging because proofs often depend on project-specific context. While iterative refinement can use compiler errors to repair failed proofs, reusing failed attempts requires careful search control: some proofs provide better starting points than others, and later revisions may degrade a partially correct proof. We propose a compiler-guided proof search framework that balances exploration and exploitation. It explores diverse starting points through dual-model generation and stagnation-triggered resampling, while exploiting promising proof states through current-best refinement guided by compiler-grounded pairwise comparison. Experiments on seven real-world Lean~4 projects from miniCTX-v2 show that our method achieves a better effectiveness--efficiency tradeoff than pass@$k$ baselines. Within the pass@$32$ budget, our method improves average pass rate by 12.8 percentage points while reducing LLM calls by 21.9\%.
\end{abstract}

\section{Introduction}
\label{sec:intro}

Large language models (LLMs) are increasingly used as agents that revise their outputs based on feedback from external environments \citep{yang2023intercode, chen2023teaching, first2023baldur}. Formal theorem proving is well suited for studying this process: in Lean~\citep{moura2021lean}, every candidate proof can be checked by a compiler, and failed attempts receive structured feedback from the compiler \citep{chen2023teaching, first2023baldur}. 

Recent neural theorem provers have made substantial progress on formal reasoning problems. 
Systems such as DeepSeek-Prover-v2~\citep{ren2025deepseek} and AlphaProof~\citep{hubert2026olympiad} show that neural provers can generate strong proof attempts, especially on olympiad-style and library-based benchmarks~\citep{zheng2021minif2f, tsoukalas2024putnambench}. However, proving theorems inside project-level Lean projects introduces additional challenges. Real-world project-level proofs often depend on local definitions, project-specific lemmas, naming conventions, and proof patterns spread across the surrounding context~\citep{tooby2025heplean}. Recent efforts on project-level Lean theorem proving~\citep{hu2024minictx,kumarappan2024leanagent, poiroux-etal-2025-rlmeval} show that even strong provers still struggle in this setting.

Most existing methods rely on independent sampling: a prover generates multiple complete proof candidates, and success is measured by pass@$k$~\citep{chen2021evaluating,ren2025deepseek}, which counts whether any candidate verifies. However, this paradigm treats the compiler as a final filter rather than a source of feedback for guiding later attempts. As a result, failed proofs are discarded even when they contain useful partial progress. In context-dependent formal proving, this partial progress may include a relevant project-specific lemma, a promising proof outline, or a nearly correct tactic sequence. This forces later samples to start largely from scratch.

Compiler-guided refinement~\citep{first2023baldur,zhou2025solving} offers a natural way to reuse failed proofs, but effective refinement is not guaranteed. Execution feedback can improve code generation and agentic problem solving~\citep{chen2021evaluating, yang2023intercode, first2023baldur}, but prior work also shows that self-correction is not reliably beneficial in general~\citep{kamoi2024can, adnan2025measuring}. Similarly, in compiler-guided proof refinement, additional revision does not necessarily improve the proof state. A revision may fix one error while introducing another, or degrade a partially correct proof. Moreover, refinement is also sensitive to the starting proof: some failed attempts are close to repairable, while others lead to stuck trajectories even after many refinements. These observations suggest an exploration--exploitation tradeoff between trying new proofs and improving existing ones. An effective prover should therefore explore diverse starting proofs, preserve useful intermediate proofs during refinement, and restart when the current path stops improving. In practice, Lean-specialized provers and general-purpose reasoning models exhibit complementary strengths on project-level problems: the former generate precise tactic sequences but underutilize surrounding project context, while the latter better leverage this context but more often produce ill-typed tactics~\citep{wang2024theoremllama}.

To this end, we propose a compiler-guided adaptive proof search framework for context-dependent Lean theorem proving. The framework combines cross-model exploration with controlled refinement. For exploration, it draws candidates from two complementary models: a Lean-specialized prover that provides tactic-level precision and a general-purpose reasoning model that can better use long project context. A compiler-grounded pairwise comparison selects the more promising candidate as the current-best proof state. For exploitation, the system refines this current-best proof using compiler feedback, but accepts a revision only when pairwise comparison indicates improvement. When refinement stops making progress, it returns to exploration and resamples candidates from both models.

Our contributions are as follows:
\begin{itemize}

    \item We introduce a hybrid proof search framework that uses complementary specialist and generalist models for exploration, and combines them with current-best refinement and stagnation-triggered resampling.

    \item We analyze proof search trajectories in context-dependent Lean theorem proving, showing that success depends strongly on the starting proof and on preserving useful intermediate proof states.

    \item We evaluate our framework on seven Lean~4 projects from miniCTX-v2 and 84 formalization problems from RLMEval-FLT3, showing that it achieves a better pass rate--LLM call tradeoff.
\end{itemize}
\noindent\textbf{Code.}
The code is available at \url{https://github.com/joeliuz6/lean_proof_search}.
\section{Related Work}

\subsection{Context-Dependent Theorem Proving}

Context-dependent theorem proving studies problems where the proof depends on surrounding project context. miniCTX~\citep{hu2024minictx} introduces a benchmark built from seven real-world Lean projects, highlighting the difficulty of theorem proving with long project-level context. RLMEval~\citep{poiroux-etal-2025-rlmeval} evaluates neural theorem proving and proof autoformalization on research-level theorems from Lean Blueprint formalization projects. LeanAgent~\citep{kumarappan2024leanagent} further studies this setting across diverse Lean repositories.
Our work targets this real-world setting and focuses on how to reuse failed proof attempts.

\subsection{Whole-Proof Generation}

Whole-proof generation produces a complete proof in one pass, rather than predicting the proof tactic by tactic. Recent formal reasoning systems have shown strong progress. AlphaProof~\citep{hubert2026olympiad} and AlphaGeometry~\citep{trinh2024solving} demonstrated that AI systems can reach medal-level performance on International Mathematical Olympiad (IMO) problems, while neural provers~\citep{wang2025kimina,ren2025deepseek,chen2025seed, lin2025goedel} have achieved strong results on olympiad-style benchmarks such as miniF2F~\citep{zheng2021minif2f} and PutnamBench~\citep{tsoukalas2024putnambench}. Other systems use verifier or compiler feedback to repair failed complete proofs~\citep{zhou2025solving,first2023baldur,lu2025adaptive, wang2026learning}, showing that failed attempts can provide useful information for later refinement. Our work builds on whole-proof generation, but focuses on search over failed proofs: we use complementary models to find diverse starting points and compiler-grounded comparison to preserve promising proof states.

\begin{figure*}[t]
    \centering
    \includegraphics[width=2.1\columnwidth]{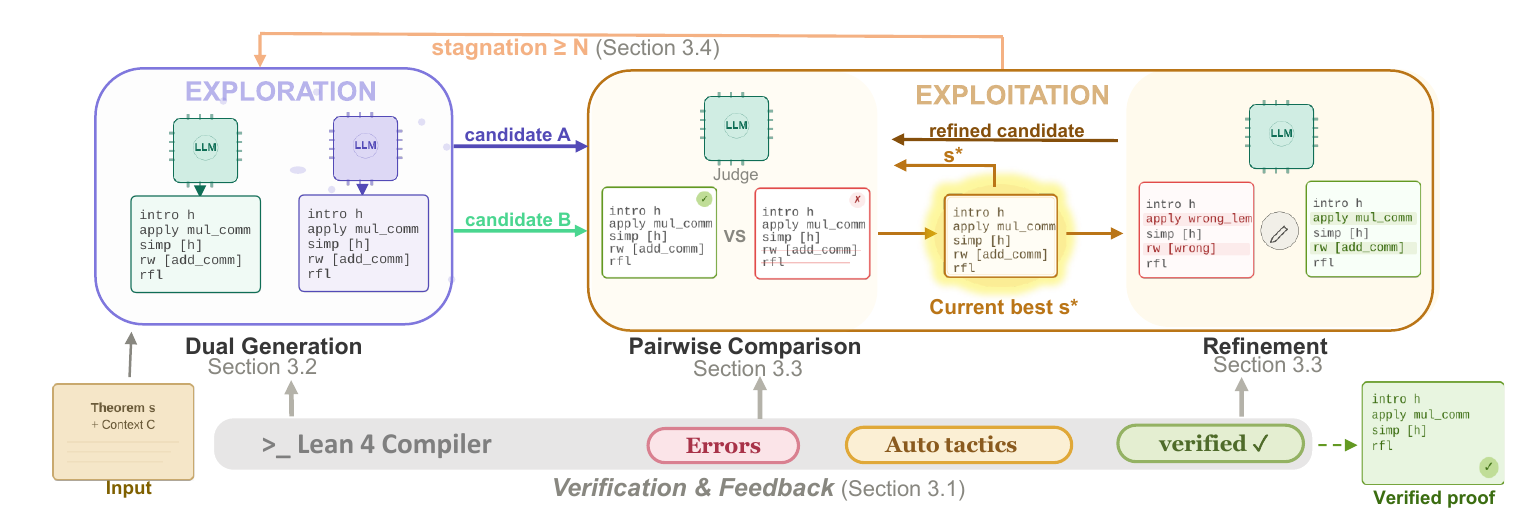}
    \caption{Overview of our framework. \textbf{Exploration} (left) generates two diverse candidates via a generalist and a specialist. \textbf{Exploitation} (right) maintains a single current-best proof $s^*$ and drives it toward verification: each repair produces a \emph{proposal}, and an LLM judge decides whether to accept it. When refinement stagnates for $N$ consecutive rounds, the system re-enters exploration and draws fresh candidates from both models. The \textbf{Verification \& Feedback} layer (bottom) provides structured error feedback, tactic suggestions and verification signals.}
    \label{fig:overview}
\end{figure*}

\subsection{Tactic Search-Based Proving}
A complementary line of work studies tactic-level proof search, where a model predicts the next tactic from the current proof state. Many systems guide this process with best-first search~\citep{xin2025bfs, wu2024internlm2, polu2020generative} or related tree-search methods~\citep{coulom2006efficient}. For example, \citet{lample2022hypertree} propose an AlphaZero-style hypertree proof-search algorithm to improve search efficiency. LeanProgress~\citep{george2025leanprogress} trains a model to predict proof progress and uses this signal to guide search. BFS-Prover~\citep{xin2025bfs} introduces a scalable best-first search framework with preference optimization over state--tactic pairs and length normalization.
Our work is different in granularity: rather than searching over individual tactics, we search over whole-proof candidates. This allows us to combine exploration from multiple proof generators with refinement of promising proof states.

\section{Compiler-Guided Adaptive Proof Search}
\label{sec:method}



\paragraph{Overview.} Given a theorem statement $s$ and its surrounding Lean~4 project context $C$, the system generates a verified proof $p$. $C$ is the source content before the target theorem in the project file. We design a compiler-guided search procedure that orchestrates two complementary off-the-shelf models with comparison-based refinement.

We view whole-proof generation as a search problem over proof candidates.
The system maintains a single \emph{current-best proof state} $s^*$ and alternates between two phases, as illustrated in Figure~\ref{fig:overview}:

\begin{itemize}
    \item \textbf{Exploration} proposes diverse starting points by drawing candidates from two models with complementary strengths: a generalist and a specialist (\S\ref{sec:dualmodel}), and re-invoking both whenever the search stagnates (\S\ref{sec:resampling}).
    \item \textbf{Exploitation} refines the current-best proof using compiler feedback, treating each repair as a proposal that may or may not replace $s^*$ (\S\ref{sec:comparerefine}).
\end{itemize}

\noindent\textbf{Pairwise comparison} serves as the search controller in the whole process: it selects the stronger initial proof during exploration, decides whether a refined proposal should replace $s^*$ during exploitation, and chooses a new starting point after resampling (\S\ref{sec:comparerefine}). 

Algorithm~\ref{alg:main} in Appendix~\ref{app:algorithm} shows the full search procedure.

\subsection{Compiler-Grounded Feedback}
\label{sec:Compiler}

We use the Lean~4 compiler not only as a verifier, but also as a source of structured error feedback and lightweight automatic tactics.

\paragraph{Structured error feedback.}
When a proof fails, we extract structured errors: the error position, and the error message, aligned to the specific tactic line that produced it. This gives models precise line-level grounding instead of file-level noise.
The error message is used by both the repair model and the pairwise comparison controller. 

\paragraph{Lightweight automatic solving.}
In our work, an \textsc{AutoSolve} module tries a small set of auto tactics: deterministic closing tactics and Lean's built-in lemma search to close trivial goals. More details can be found in Appendix~\ref{app:impl}.
Proofs solved this way exit the search without any model call, saving budget for harder goals.

\subsection{Exploration: Dual-Model Candidate Generation}
\label{sec:dualmodel}

A single model often produces similar errors across repeated failed samples. Therefore, we generate one candidate from each of two complementary models to add meaningful diversity.

\paragraph{Generalist.}
We use a strong general-purpose reasoning model as the generalist, e.g., GPT-5~\citep{singh2025openai}. Its long-context capacity makes it suitable for using project-specific definitions and dependencies, although it may still produce invalid Lean syntax or tactic usage~\citep{wang2024theoremllama}.

\paragraph{Specialist.}
We use a Lean-specialized prover~\citep{wang2025kimina,ren2025deepseek} as the specialist, e.g., Deepseek Prover~\citep{ren2025deepseek}. Because it is trained on large-scale Lean proof data, it is better aligned with Lean syntax, tactics, and common proof patterns, but may be less effective at using long project-level context.

\medskip
\noindent The two models often perform in different ways: the generalist is stronger at context understanding, while the specialist is stronger at formal proof syntax. As a result, they tend to produce different kinds of proof attempts rather than repeated variations of the same mistake. In our method, each candidate is verified immediately after generation; if either proof succeeds, the system returns without refinement. Otherwise, pairwise comparison selects the more promising proof as the initial $s^*$.

\subsection{Exploitation: Comparison-Based Current-Best Refinement}
\label{sec:comparerefine}

\paragraph{Why refinement needs search control.}
Standard iterative refinement~\cite{first2023baldur} treats each revision as an unconditional state update:
\begin{equation}
    s_{t+1} = \textsc{Refine}(s_t, e_t),
    \label{eq:naive}
\end{equation}
using compiler error feedback $e_t$, and overwriting $s_t$ regardless of whether $s_{t+1}$ is actually better. However, refinement is not always an improvement. A revised proof may fix one error but introduce new ones, or move away from a previously promising strategy. While a single bad revision may not matter much, these mistakes can accumulate over many rounds and gradually push the search away from a correct proof.

\begin{figure}[t]
    \centering
    \includegraphics[width=\columnwidth]{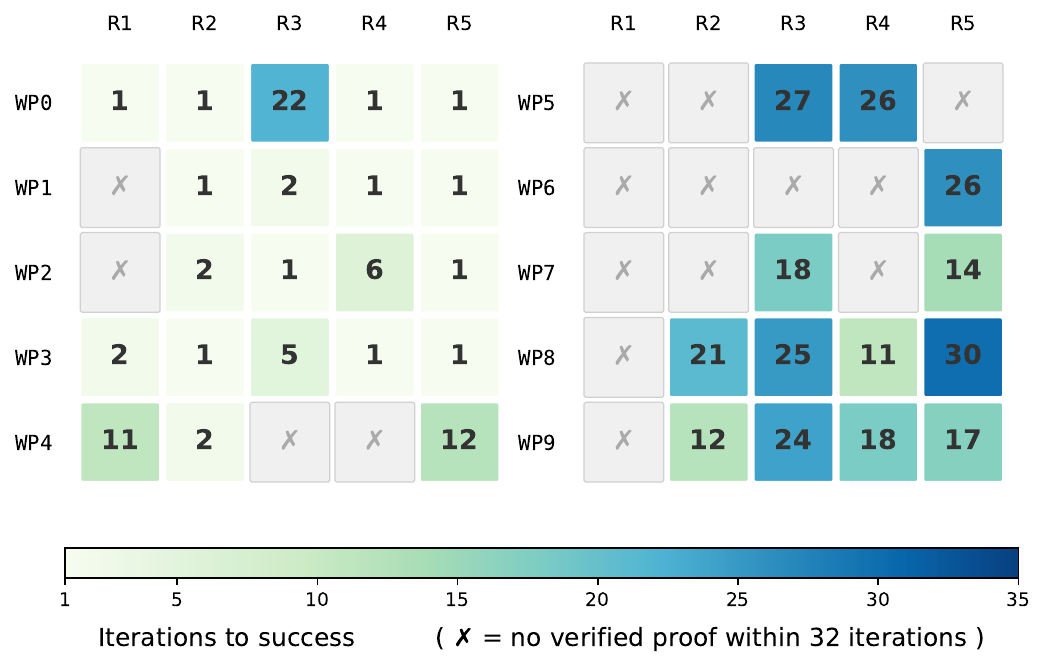}
    \caption{\textbf{Empirical refinement results of one theorem with different initial attempts.} One theorem, 10 wrong proofs (WP) as starting points (rows), each evaluated with 5 independent refinement runs (columns) with a 32-iteration budget. Cells show iterations to success; $\times$ marks failure. Theorem: \texttt{Set\_EAnnulus\_oc\_subset\_co} in \textsc{carleson}. More examples can be found in Figure~\ref{fig:heatmap-set} and Figure~\ref{fig:heatmap-elpnorm} in Appendix~\ref{app:refinement-randomness}.}
    \label{fig:motivation}
\end{figure}

Figure~\ref{fig:motivation} illustrates two important patterns. First, the choice of starting proof matters a lot. For the same theorem, some wrong proofs (WP0--WP3) can usually be repaired within a few refinement steps, while others (WP5, WP6) rarely succeed under the same budget. We further verify in Appendix~\ref{app:refinement-randomness} that this effect is not only due to sampling randomness: variation across different starting proofs is larger than variation across repeated runs from the same starting proof. Second, the effect of refinement is not consistent across different runs. Even when starting from the same wrong proof, different runs can follow different trajectories: some quickly reach a correct proof, while others drift away and fail.

\paragraph{Propose-then-accept refinement.}
Instead of always replacing the current proof with a new revision, we use a propose-then-accept strategy.
At each step, the repair model generates a revised proof $\tilde{s}_{t+1}$ from the current-best proof $s^*_t$ and its compiler errors $e^*_t$:
\begin{align}
    \tilde{s}_{t+1} &= \textsc{Refine}(s^*_t, e^*_t), \\
    s^*_{t+1} &= \textsc{Better}(s^*_t,\ \tilde{s}_{t+1}).
\end{align}
If the new proof verifies, the search stops. Otherwise, $\textsc{Better}$ decides whether the new proof is more promising than the current-best proof.

\begin{figure*}[t]
    \centering
    \includegraphics[width=\textwidth]{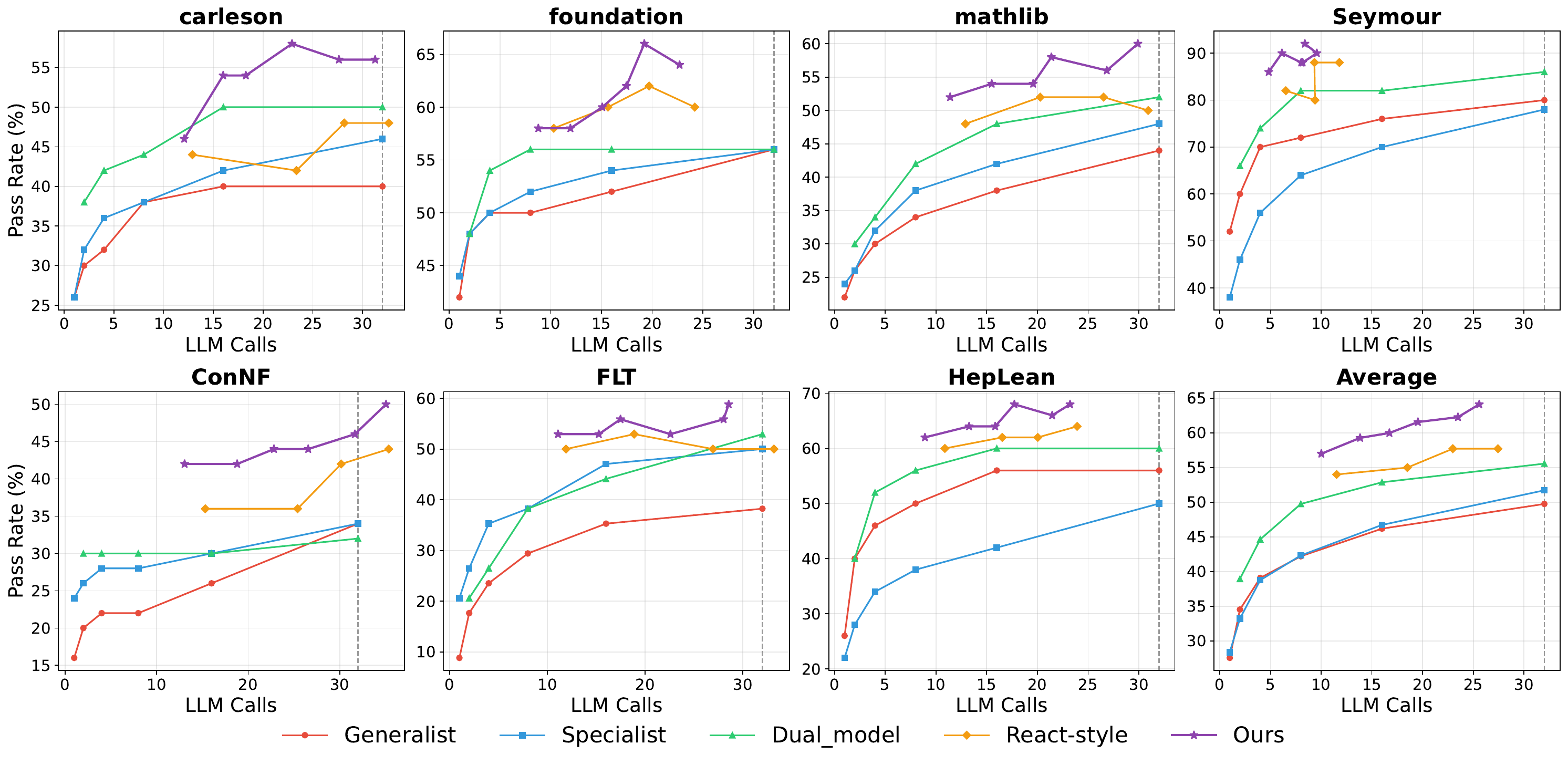}
    \caption{Pass rate (\%) \emph{vs.} LLM calls across seven Lean~4 projects from \texttt{miniCTX-v2}. Gray dashed line is pass@32. }
    \label{fig:main_results}
\end{figure*}

\paragraph{Pairwise comparison.}
$\textsc{Better}$ is an LLM judge that compares two proofs together with their compiler errors. It chooses the proof that looks more likely to succeed after future refinement, based on factors such as current progress, error severity, and overall proof strategy. If the new proof is accepted, it becomes the new current-best proof $s^*$. Otherwise, the system keeps the old one. Rejected proposals increase a stagnation counter $n_{\text{stag}}$, while accepted proposals reset it to zero.

\subsection{Returning to Exploration: Stagnation-Aware Resampling}
\label{sec:resampling}

Pairwise comparison helps preserve promising proof states, but it cannot fix a fundamentally bad proof strategy. Sometimes the current proof gets stuck in a region where local refinement no longer helps: new revisions keep getting rejected for several iterations, and there are increasing errors in refinement. Resampling serves as a fallback mechanism in this situation.

\paragraph{Trigger and restart.}
When the stagnation counter $n_{\text{stag}}$ reaches a threshold $N$, the system returns to exploration. It generates fresh candidates from both models, verifies them, and uses $\textsc{Better}$ to choose a new current-best proof $s^*$. Refinement then continues from this new starting point. We resample from both models instead of only the model that produced the stagnated proof. This allows different proof strategies and helps avoid repeatedly exploring the same bad region of proof space.



\begin{figure}[t]
  \centering
  \includegraphics[width=0.48\textwidth]{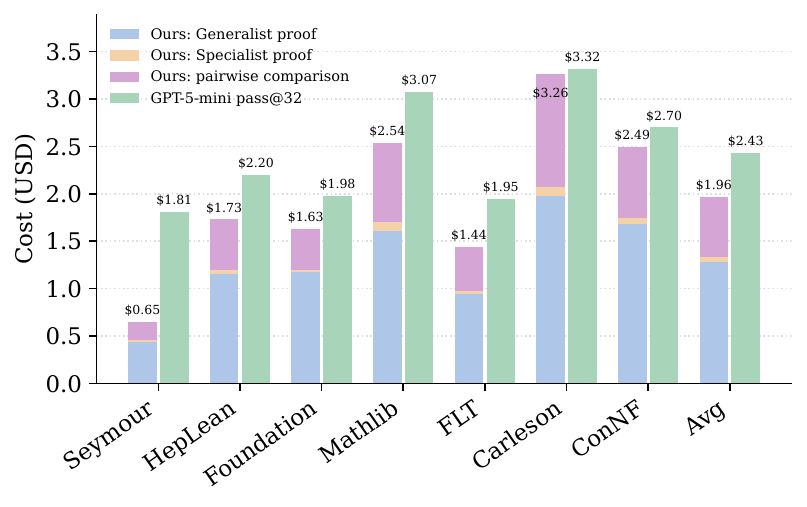}
  \caption{Total LLM cost breakdown per project under 32 average LLM calls. Specialist DeepSeek-Prover-v2-7B is self-hosted; its cost is estimated at the generalist's per-token price (higher than the specialist) for a conservative comparison.}
  \label{fig:cost_break}
\end{figure}

\begin{table*}
  \centering
  \setlength{\tabcolsep}{4pt}
  \resizebox{0.8\textwidth}{!}{%
  \begin{tabular}{@{} l *{7}{c} @{\hskip 12pt} c @{}}
    \toprule
    \multirow{2}{*}{\textbf{Method}}
      & \multicolumn{7}{c}{\textbf{Pass rate (\%) per project}}
      & \multirow{2}{*}{\textbf{Avg.}} \\
    \cmidrule(lr){2-8}
      & Carleson & ConNF & FLT & Foundation & HepLean & Mathlib & Seymour & \\
    \midrule
    Gemini-2.5-Flash
      & 44.00 & 28.00 & 35.29 & 52.00 & 38.00 & 40.00 & 72.00 & 44.18 \\[2pt]
    DeepSeek-Prover-v2
      & 42.00 & 30.00 & 47.06 & 54.00 & 42.00 & 42.00 & 70.00 & 46.72 \\[2pt]
    \midrule
    \textbf{Ours}
      & \textbf{48.00} & \textbf{34.00} & \textbf{50.00}
      & \textbf{60.00} & \textbf{54.00} & \textbf{46.00}
      & \textbf{78.00} & \textbf{52.86} \\
    \quad {\footnotesize\textcolor{gray}{avg calls}}
      & {\footnotesize\textcolor{gray}{14.86}}
      & {\footnotesize\textcolor{gray}{16.46}}
      & {\footnotesize\textcolor{gray}{15.03}}
      & {\footnotesize\textcolor{gray}{12.28}}
      & {\footnotesize\textcolor{gray}{14.92}}
      & {\footnotesize\textcolor{gray}{16.12}}
      & {\footnotesize\textcolor{gray}{8.98}}
      & {\footnotesize\textcolor{gray}{14.09}} \\
    \bottomrule
  \end{tabular}%
  }
  \caption{Pass rate (\%) on \texttt{miniCTX-v2} across seven projects with \textbf{Gemini-2.5-Flash} as the Generalist. Results of full proof generation are pass@16.
           Our method additionally reports average LLM calls per theorem (\textcolor{gray}{gray}).}
  \label{tab:gemini}
\end{table*}

\section{Experimental Setup}
\label{sec:setup}

\subsection{Benchmark}

\paragraph{miniCTX-v2.}
Our main benchmark is miniCTX-v2~\citep{hu2024minictx}, a context-dependent theorem proving benchmark built from seven real-world Lean~4 projects. 
The projects differ in mathematical domain, proof style and dependency structure, making miniCTX-v2 a useful benchmark for evaluating whether proof search methods generalize across different Lean environments. We report our results on the test split.

\paragraph{RLMEval-FLT3.}
We also evaluate our method on the FLT3 subset of RLMEval~\citep{poiroux-etal-2025-rlmeval}, which studies real-world formalization problems.
FLT3 contains 84 theorems from a Lean~4 formalization of Fermat's Last Theorem for $n=3$, each paired with a natural language proof. 
Following RLMEval, we evaluate in the \emph{easy} setting, where all project lemmas are available during proof generation.

\subsection{Models}

Unless otherwise noted, we choose \textbf{DeepSeek-Prover-v2-7B}~\citep{ren2025deepseek} as the specialist and \textbf{GPT-5-mini}~\citep{singh2025openai} as the generalist, which also serves as the repair model and pairwise judge.

\subsection{Baselines}

\paragraph{Single-model generation (Generalist / Specialist).}
We follow the standard pass@$k$~\cite{chen2021evaluating} protocol: each model draws $k$ independent candidates and succeeds if any verifies, consistent with prior evaluations~\citep{ren2025deepseek}.
 
\paragraph{Dual-model generation (Dual\_Model).}
Both models sample independently, and their outputs are combined into a single candidate set.
At a budget of $2k$ calls, each model contributes $k$ samples; it succeeds if any candidate verifies.
 
\paragraph{Agentic baseline (ReAct-Style).}
We implement a ReAct-style agent~\citep{yao2022react} under the 
same model access, AutoSolve, compiler feedback, and verification harness as 
our method. The agent maintains a bank of proof candidates, each 
paired with its compiler errors. At each turn, an LLM controller 
(GPT-5-mini) observes the current proof bank and recent action 
history, then selects one of three actions: (1)~sample from generalist, (2)~sample from specialist, or (3)~refine an existing candidate by 
specifying its index.

\begin{figure}[t]
  \centering
  \includegraphics[width=0.47\textwidth]{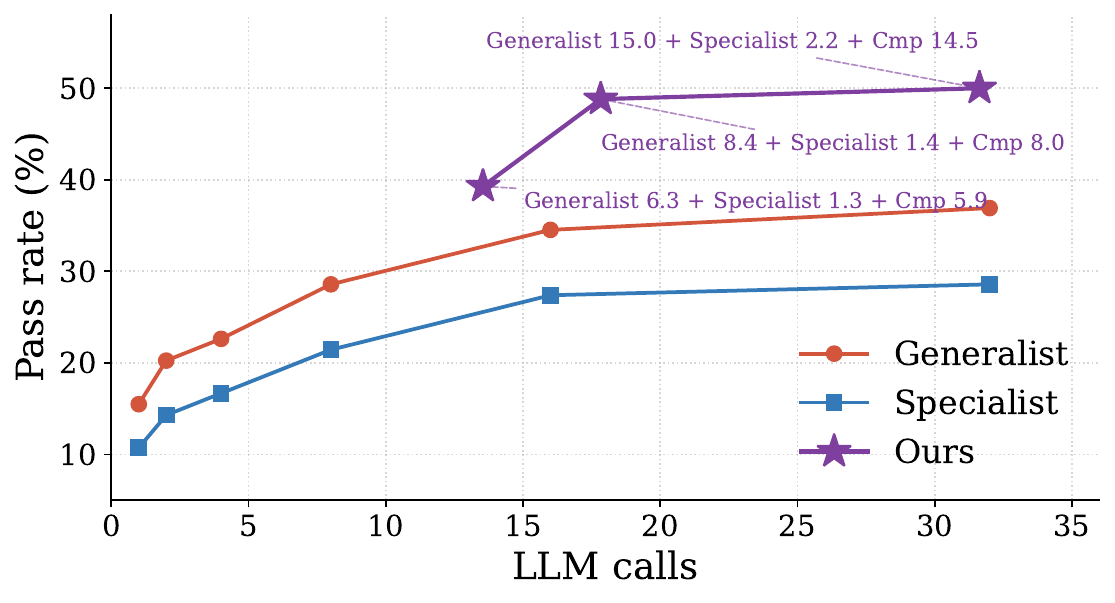}
  \caption{Pass rate (\%) \emph{vs.} LLM calls on \texttt{RLMEval-FLT3} with informal proof guidance. Text labels show average calls from the generalist, specialist, and pairwise comparison (Cmp).}
  \label{fig:flt3}
\end{figure}
\subsection{Metrics}
\label{sec:metrics}
 
We report \textbf{pass rate} (fraction of theorems verified) against \textbf{average LLM calls per theorem}. The pass-rate vs.\ LLM-call curve is our main evaluation, capturing the effectiveness--efficiency tradeoff. We additionally break down cost in dollars (Figure~\ref{fig:cost_break}).

\subsection{Implementation Details}
\begin{figure*}[t]
  \centering
  \begin{subfigure}[b]{0.38\textwidth}
    \centering
    \includegraphics[width=\textwidth]{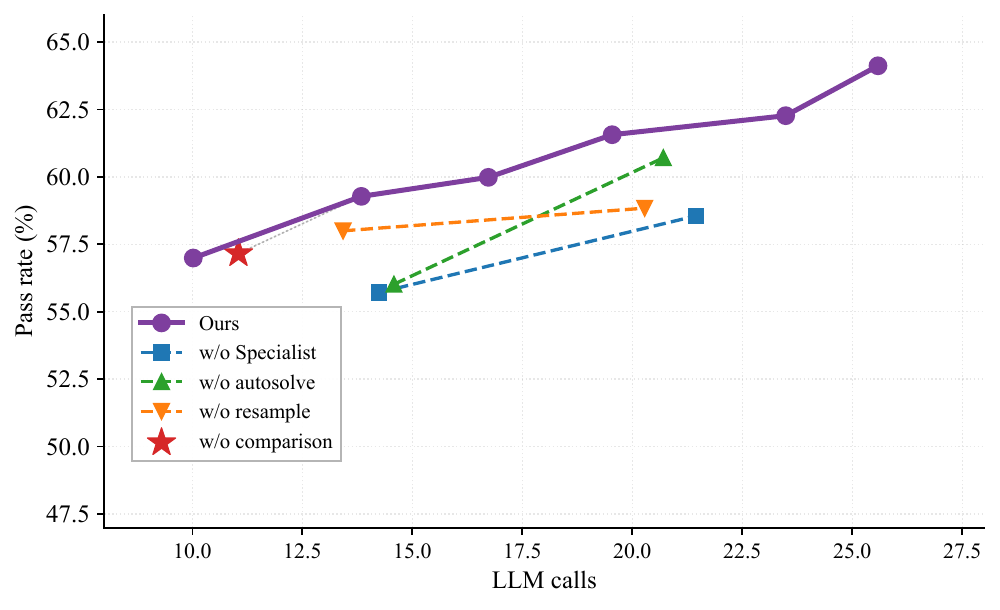}
    \caption{Ablation study of different components.}
    \label{fig:ablation_main}
  \end{subfigure}
  \hfill
  \begin{subfigure}[b]{0.6\textwidth}
    \centering
    \includegraphics[width=\textwidth]{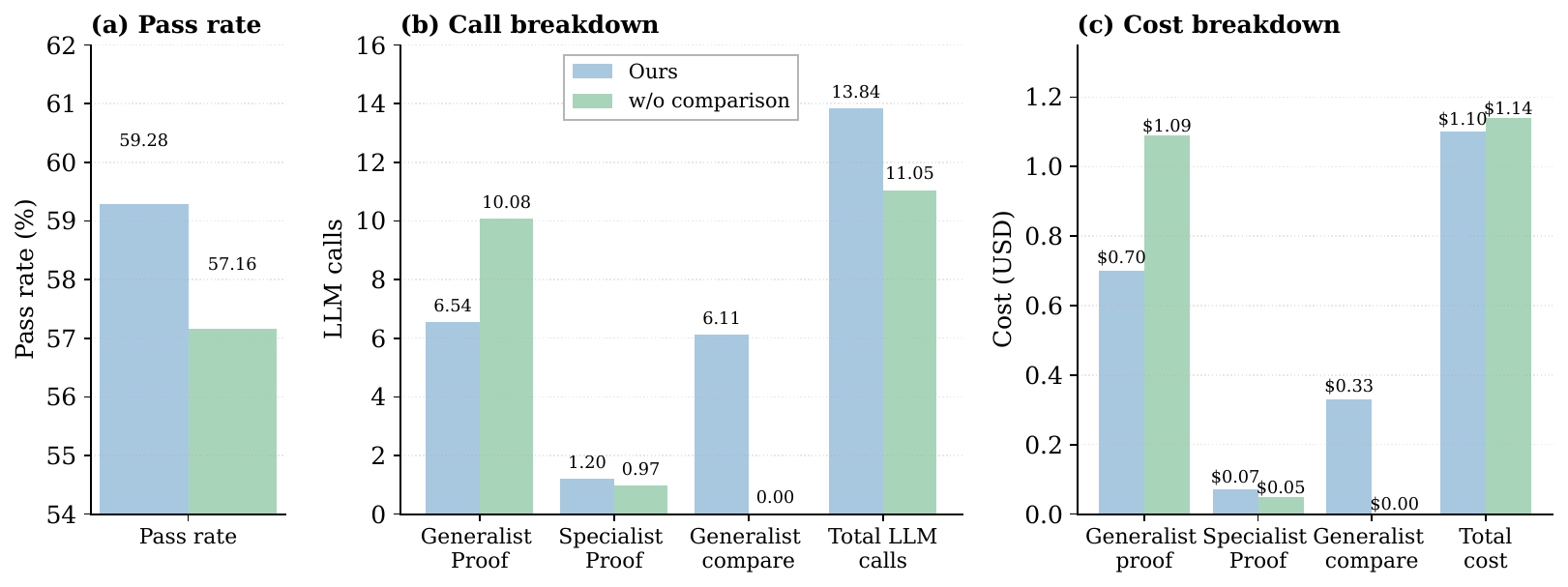}
    \caption{Effect of pairwise comparison on call allocation and cost.}
    \label{fig:ablation_detail}
  \end{subfigure}
  \caption{Ablation study results.}
  \label{fig:ablation}
\end{figure*}

\begin{figure}[t]
    \centering
    \includegraphics[width=0.9\columnwidth]{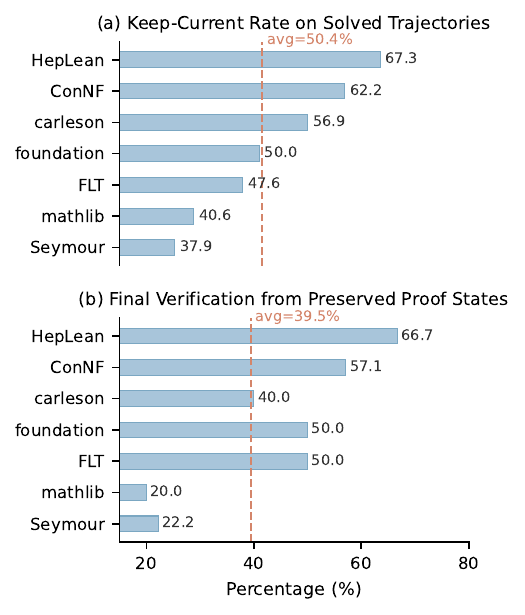}
    \caption{Pairwise comparison behavior on solved refinement trajectories.
    (a)~Keep-current rate macro-averaged over trajectories during successful refinement.
    (b)~Final verification from preserved proof states. Dashed lines are pooled over all comparison steps.}
    \label{fig:reject_rate_thm}
\end{figure}
Our method interacts with Lean~4 through the \texttt{lean-interact} library~\citep{leaninteract}. 
For a fair comparison, we use low reasoning effort for the generalist model. For the specialist, we set temperature~1 with top-$p$~0.95 to encourage diverse proofs. Full details are in Appendix~\ref{app:impl}.

\section{Results}
\label{sec:results}

\subsection{Main Results on miniCTX-v2}
\label{sec:main_results}

We evaluate all methods by plotting pass rate against the average number of LLM calls per theorem.
For our method, we vary the max refinement budget $K \in \{8, 12, 16, 20, 24, 28\}$ and set the stagnation threshold to $N=\lceil K/4 \rceil$ following the analysis in Table~\ref{tab:threshold}.
For each baseline, we vary its call budget up to an average budget of 32 calls.

\paragraph{Our method achieves the best effectiveness--efficiency tradeoff.}
As shown in Figure~\ref{fig:main_results}, our method is consistently at or near the top of the pass rate curve across projects. Within the pass@32 budget (left to the gray line), our method improves the average pass rate by 12.8 percentage points over the average pass@32 performance of the two base models, while reducing LLM calls by 21.9\%.
 
\paragraph{Our method balances exploration and exploitation.}
Dual-model generation and stagnation-triggered resampling serve as exploration: they produce diverse proof candidates. As shown in Figure~\ref{fig:main_results}, dual-model generation already outperforms either single model. Compiler-guided refinement serves as exploitation: it improves the current-best proof using structured error feedback. At higher call budgets, the dual-model generation flattens (e.g., \textsc{carleson} plateaus at 0.50 after about 16 calls), while our method continues to improve.
 
\paragraph{Structured control outperforms open-ended action selection.}
As shown in Figure~\ref{fig:main_results}, our method outperforms the ReAct-style baseline on every project, despite a simpler controller. This suggests that, reliable search benefits from an explicit exploration--exploitation schedule rather than leaving search decisions entirely to an LLM controller.

\paragraph{Cost.}
Figure~\ref{fig:cost_break} and Figure~\ref{fig:call_break} (Appendix~\ref{sec:call_breakdown}) break down the per-project LLM cost and the average number of calls per theorem for the points under 32 average LLM calls (left to the gray dashed line in Figure~\ref{fig:main_results}).
We count three types of LLM calls: generalist proof calls, including initial proof generation and refinement; specialist proof calls; and pairwise comparison calls.
On average, our method spends \$1.96, compared with \$2.43 for GPT-5-mini pass@32, while achieving a higher pass rate. Although comparison calls occur at a similar frequency to generalist proof calls, they are cheaper due to short decision output. See Appendix~\ref{sec:call_breakdown} for more detailed token comparison.

\subsection{Results on RLMEval-FLT3}
\label{sec:flt3}

We also test on RLMEval-FLT3, where each theorem comes with an informal proof. We provide the informal proof to the generation, refinement, and pairwise comparison modules. As Figure~\ref{fig:flt3} shows, our method remains effective in this setting: at just 13.5 calls per theorem, it already reaches 39.3\% pass rate, higher than both Generalist pass@32 (36.9\%) and Specialist pass@32 (28.6\%).

\subsection{Robustness to Model Choice}
\label{sec:robustness}

We test whether our gains depend on a specific model pairing by replacing the generalist with Gemini-2.5-Flash and rerunning the full pipeline. As Table~\ref{tab:gemini} shows, our method outperforms both base models on every project, reaching 52.86\% average pass rate under a budget of 16 calls. Results with an additional generalist-specialist pair in Appendix~\ref{app:goedel} further confirm this finding.

\section{Analysis}
\label{sec:analysis}

\subsection{Ablation Study}
\label{sec:ablation}
\begin{table}[t]
  \centering

  \small
  \begin{tabular}{@{} l c c @{}}
    \toprule
    \textbf{Component} & \textbf{Fraction (\%)} & \textbf{Subtotal (\%)} \\
    \midrule
    AutoSolve            & 28.36 & 28.36 \\
    \midrule
    Init (Generalist)    & 30.34 & \\
    Init (Specialist)    &  9.44 & 39.78 \\
    \midrule
    Resample (Generalist)&  0.49 & \\
    Resample (Specialist)&  1.01 &  1.50 \\
    \midrule
    Refine (Generalist)  & 18.91 & \\
    Refine (Specialist)  & 11.46 & 30.37 \\
    \bottomrule
  \end{tabular}
  \caption{Fraction of theorems proved by each component.}
  \label{tab:success_source}
\end{table}
Figure~\ref{fig:ablation}(a) shows the pass rate--LLM calls tradeoff for four ablations: w/o specialist, w/o autosolve, w/o resample, and w/o comparison. In the w/o comparison setting, pairwise comparison is replaced by a random selector. The ablation shows that the gain does not come from more model calls alone; it comes from how the calls are allocated across generation, refinement, comparison, and restart.

\paragraph{Every component contributes to performance.}
Removing any single component shifts the curve downward, indicating that specialist generation, autosolve, resampling, and pairwise comparison each improve the search process. Since w/o comparison removes comparison calls by design, we further analyze its call allocation and cost in Figure~\ref{fig:ablation}(b).

\paragraph{Pairwise comparison improves proof state selection.}
Without comparison, the system relies more on proof generation: generalist proof calls increase from 6.54 to 10.08 per theorem. Pairwise comparison improves proof state selection in two ways: it chooses a better starting point after initial dual-model generation, and it preserves stronger intermediate states during refinement. This prevents useful proof trajectories from being overwritten by weaker proposals. As a result, our method achieves a higher pass rate (59.28\% vs.\ 57.16\%) with slightly lower total cost (\$1.10 vs.\ \$1.14).
\subsection{Proof Search Trajectory Analysis}
\label{sec:judge_behavior}


\paragraph{Refinement proposals are frequently rejected.}
Figure~\ref{fig:reject_rate_thm}(a) shows that, even on successful trajectories, about half of refinement proposals are rejected on average ($50.4\%$). This suggests that refinement does not consistently move toward a correct proof: new revisions often introduce new errors or move away from a previously promising direction. Without comparison, these unstable updates would overwrite the current proof state at every step.

\paragraph{Successful proofs often come from preserved states.}
Figure~\ref{fig:reject_rate_thm}(b) shows that among solved trajectories whose final verification follows a pairwise comparison step, $39.5\%$ are verified after the comparison chooses the existing current-best proof state rather than the newest refinement proposal. This means that these successes come from a state preserved by comparison, not from immediately accepting the latest refinement.

Together, these results suggest that successful refinement is not just about generating new revisions, but also about preserving useful proof states.
\subsection{Complementary Sources of Success}
\label{sec:success_source}

\begin{table}[t]
\centering
\resizebox{\columnwidth}{!}{
\begin{tabular}{lccccc}
\toprule
& $K/12$ & $K/6$ & $K/4$ & $K/3$ & $K/2$ \\
\midrule
Pass Rate (\%) & 48.00 & 56.00 & \textbf{60.00} & 54.00 & 60.00 \\
LLM Calls & 30.24 & 27.12 & \textbf{24.42} & 27.60 & 27.76 \\
\bottomrule
\end{tabular}
}
\caption{Effect of stagnation threshold $N$ on mathlib-valid.}
\label{tab:threshold}
\end{table}

Table~\ref{tab:success_source} decomposes verified proofs by the stage where they are first found. The key observation is that our framework succeeds through complementary search stages rather than a single source of success. 
The generalist and specialist contribute different solved cases, consistent with their different strengths. Beyond initial generation, refinement accounts for 30.37\% of solved theorems. This shows that failed proofs often contain recoverable partial progress and that post-failure search is essential. AutoSolve closes simple goals at zero LLM cost to improve efficiency. Appendix~\ref{sec:no-autosolve} further shows that our method still outperforms the baselines when AutoSolve is disabled. Resampling rarely solves proofs directly, but serves as a fallback mechanism
when refinement stagnates, providing a new starting point for subsequent refinement. Appendix~\ref{sec:resampling-example} gives an example how resampling helps later refinement.

\subsection{Sensitivity to Stagnation Threshold}
\label{sec:sensitivity}

\begin{figure*}[t]
\centering

\begin{minipage}[t]{0.47\textwidth}
\vspace{0pt}
\textbf{Relevant source context:}
\begin{lstlisting}[basicstyle=\ttfamily\scriptsize, breaklines=true]
lemma z_spec : S.a + eta ^ 2 * S.b = lambda * S.z := ...

lemma lambda_not_dvd_y : not lambda | S.y := by
  intro h
  replace h := mul_dvd_mul_left (eta - 1) h
  rw [<- y_spec] at h
  rw [<- pow_two] at h
  exact S.lambda_sq_not_dvd_a_add_eta_mul_b h
\end{lstlisting}
\end{minipage}
\hfill
\begin{minipage}[t]{0.47\textwidth}
\vspace{0pt}
\textbf{Target theorem:}
\begin{lstlisting}[basicstyle=\ttfamily\scriptsize, breaklines=true]
lemma lambda_not_dvd_z : not lambda | S.z := by
\end{lstlisting}

\vspace{-1mm}
\textbf{Generated proof:}
\begin{lstlisting}[basicstyle=\ttfamily\scriptsize, breaklines=true]
  intro h
  replace h := mul_dvd_mul_left (eta - 1) h
  rw [<- z_spec] at h
  rw [<- pow_two] at h
  exact S.lambda_sq_not_dvd_a_add_eta_sq_mul_b h
\end{lstlisting}
\end{minipage}

\caption{
Example of context usage. The generated proof explicitly uses \texttt{z\_spec} and also follows the proof structure of the nearby theorem \texttt{lambda\_not\_dvd\_y}, while using the corresponding facts for \texttt{S.z}.
}
\label{fig:context-use-example2}
\end{figure*}

Table~\ref{tab:threshold} shows that too small $N$ triggers premature restarts, wasting calls on unnecessary resampling, while too large $N$ delays necessary restarts, spending calls on exhausted trajectories.
$N = K/4$ achieves the best balance, reaching $60.00\%$ pass rate with $24.42$ calls.

\subsection{Context Usage and Trade-offs}

We use the full context in our method, which makes all local definitions, lemmas, and proof examples available to the model without an additional retrieval component. Figure~\ref{fig:context-use-example2} shows how the source context is used. Appendix~\ref{sec:explicit-context} provides a broader analysis of how different stages of our method use source context. The main drawbacks of full context are higher input cost and potentially irrelevant context. Retrieval could reduce context length, but may fail to retrieve a necessary local fact or useful proof. Online learning could instead adapt the model to project-specific information, but requires additional training and updates. These alternatives are complementary to our framework, since the search procedure does not depend on how the context is provided.


\section{Conclusion}

This work studies adaptive proof search for context-dependent theorem proving. Our framework combines exploration from dual-model generation with exploitation from current-best refinement, allowing it to preserve and improve better proof states. More broadly, our results suggest that whole-proof generation can be extended from independent sampling to adaptive proof search. Rather than only training stronger provers, future provers may benefit from compiler-grounded process signals that can help decide which proof states to keep, refine, or abandon.

\section*{Limitations}

\paragraph{Extending complementary generation and refinement.}
Our framework uses two models for initial proof generation and a generalist model for refinement. Future work could explore larger or more specialized model pools, and study how different models should be assigned to generation, refinement, comparison, and resampling.

\paragraph{Limited use of tactic-level search.}
Our method works at the whole-proof level. 
It generates, refines, and compares complete proofs, but does not explicitly search over tactic-level proof states. 
Future work could combine our trajectory-level control with tactic-based search.

\paragraph{Reliance on pretrained inference-time models.}
Our framework uses pretrained LLMs at inference time. 
We do not train models to improve through multi-turn interaction with Lean feedback. 
Future work could explore reinforcement learning or other training methods for verifier-guided interaction.

\paragraph{Toward verifier-grounded process rewards.}
Our comparison uses an LLM judge to decide whether to keep or replace a proof. This is flexible, but it may not always reflect true proof progress. An LLM judge is a practical starting point for proof selection. Learning verifier-grounded process reward models, or using an independent judge, could provide more reliable guidance for keeping, refining, or restarting proof trajectories.

\section*{Acknowledgments}
We thank the anonymous reviewers for their helpful suggestions. This work was supported in part by computational resources provided through the National Artificial Intelligence Research Resource (NAIRR) Pilot under Award NAIRR250254.

\bibliography{latex/my_bib}

\clearpage

\appendix

\section{Algorithm}
\label{app:algorithm}

Algorithm~\ref{alg:main} shows the full search procedure.
During refinement, the current-best proof $s^*$ changes only when a new proposal is accepted (lines~\ref{line:accept}--\ref{line:reject}).
During resampling, $s^*$ is replaced by a fresh proof generated during exploration (lines~\ref{line:resample_start}--\ref{line:resample_end}).
The threshold $N$ controls how often the search switches between exploration and refinement: smaller $N$ encourages more exploration, while larger $N$ spends more budget on refinement.

\begin{algorithm}[h]
\caption{Compiler-Grounded Refinement Search with Dual-Model Exploration}
\label{alg:main}
\begin{algorithmic}[1]
\REQUIRE Theorem $s$, context $C$, max iters $K$, stagnation threshold $N$
\STATE $\mathcal{E} \leftarrow \textsc{ValidateContext}(C)$
\IF{$\textsc{AutoSolve}(s, \mathcal{E})$ succeeds}
    \RETURN proof
\ENDIF
\STATE $p_{\text{gen}} \leftarrow \textsc{Generalist}(C, s)$
\IF{$\textsc{Verify}(p_{\text{gen}})$}
    \RETURN $p_{\text{gen}}$
\ENDIF
\STATE $p_{\text{spec}} \leftarrow \textsc{Specialist}(C, s)$
\IF{$\textsc{Verify}(p_{\text{spec}})$}
    \RETURN $p_{\text{spec}}$
\ENDIF
\STATE $s^* \leftarrow \textsc{Better}(p_{\text{gen}}, p_{\text{spec}})$
\STATE $n_{\text{stag}} \leftarrow 0$
\FOR{$k = 1$ to $K$}
    \IF{$n_{\text{stag}} \geq N$} \label{line:resample_start}
        \STATE $r_1 \leftarrow \textsc{Model}_{\text{winner}}(C, s)$
        \STATE $r_2 \leftarrow \textsc{Model}_{\text{other}}(C, s)$
        \IF{$\textsc{Verify}(r_1)$ \textbf{or} $\textsc{Verify}(r_2)$}
            \RETURN verified proof
        \ENDIF
        \STATE $s^* \leftarrow \textsc{Better}(r_1, r_2)$;\ $n_{\text{stag}} \leftarrow 0$ \label{line:resample_end}
        \STATE \textbf{continue}
    \ENDIF
    \STATE $\tilde{s} \leftarrow \textsc{Refine}(C, s^*, \textsc{Errors}(s^*))$
    \IF{$\textsc{Verify}(\tilde{s})$}
        \RETURN $\tilde{s}$
    \ENDIF
    \IF{$\textsc{Better}(s^*, \tilde{s}) = \tilde{s}$} \label{line:accept}
        \STATE $s^* \leftarrow \tilde{s}$;\ $n_{\text{stag}} \leftarrow 0$
    \ELSE \label{line:reject}
        \STATE $n_{\text{stag}} \leftarrow n_{\text{stag}} + 1$
    \ENDIF
\ENDFOR
\RETURN failure
\end{algorithmic}
\end{algorithm}

\begin{figure}[t]
    \centering
    \includegraphics[width=\columnwidth]{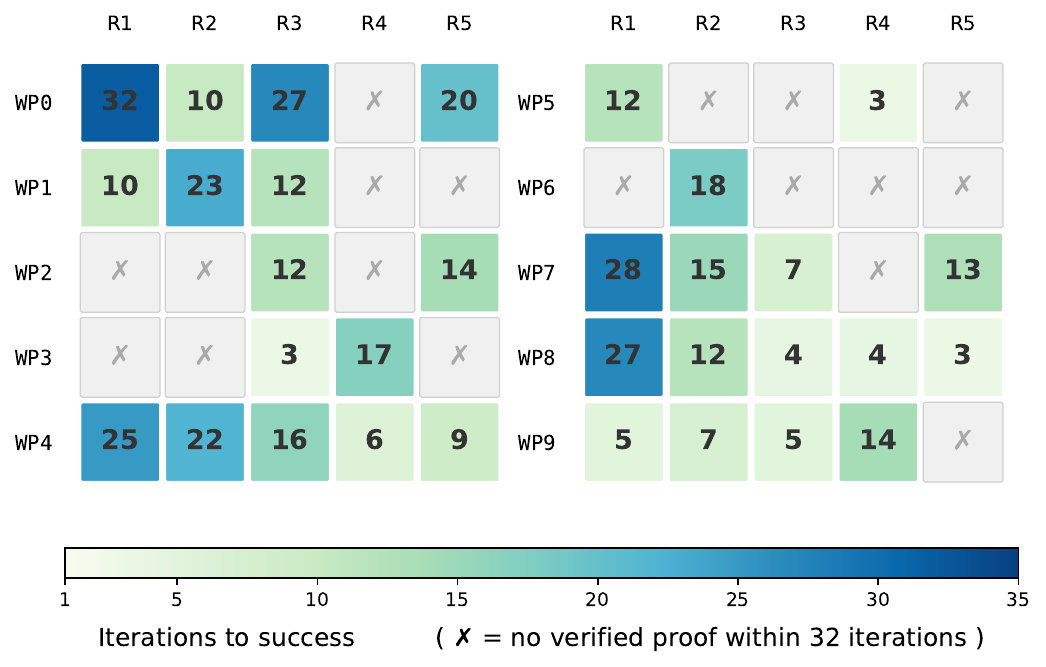}
    \caption{Refinement heatmap for \texttt{\_isBigO\_deriv\_ofReal\_cpow\_const\_atTop} in mathlib.}
    \label{fig:heatmap-set}
\end{figure}

\begin{figure}[t]
    \centering
    \includegraphics[width=\columnwidth]{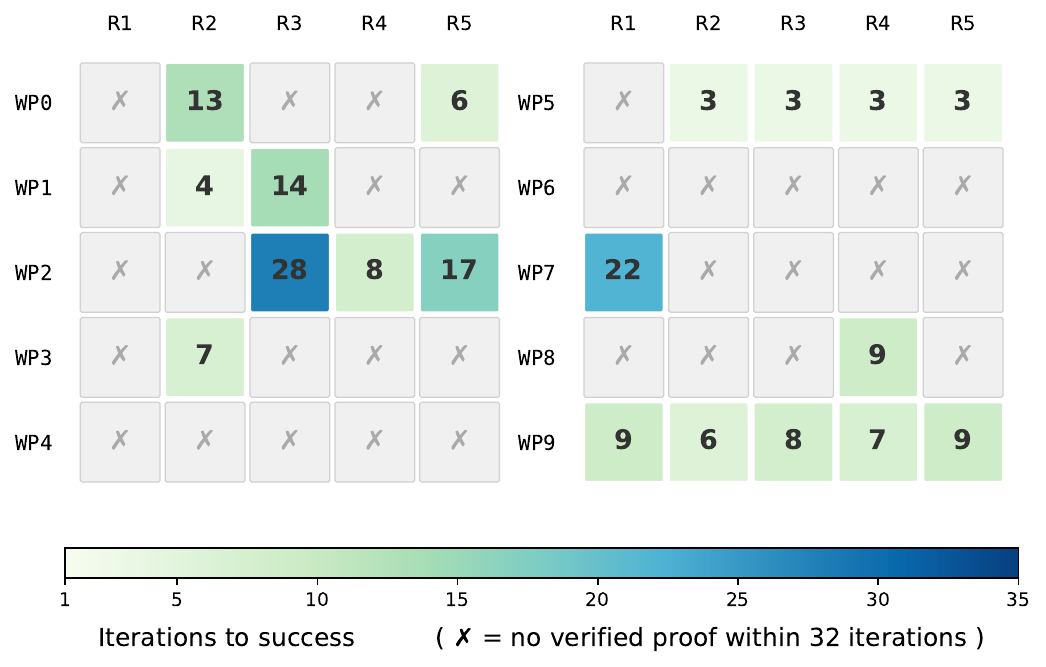}
    \caption{Refinement heatmap for \texttt{eLpNorm\_toReal\_le} in Carleson.}
    \label{fig:heatmap-elpnorm}
\end{figure}

\section{Additional Analysis}
\subsection{Analysis of Randomness in Iterative Refinement}
\label{app:refinement-randomness}

To understand how much the choice of starting wrong proof affects refinement outcomes versus inherent stochasticity, we run five independent refinement runs for each wrong proof and measure two sources of variation:

\begin{itemize}
    \item \textbf{Intra-WP std}: the standard deviation of success iterations across multiple reruns of the \emph{same} wrong proof, capturing run-to-run randomness.
    \item \textbf{Inter-WP std}: the standard deviation of success iterations across \emph{different} wrong proofs within the same theorem, capturing the effect of starting point selection.
\end{itemize}
Failed runs (no verified proof within 32 iterations) are assigned a value of 32.

Figures~\ref{fig:heatmap-set} and~\ref{fig:heatmap-elpnorm} visualize the iteration counts for two theorems. Each row corresponds to a wrong proof starting point, and each column to an independent rerun. The prevalence of entire rows marked with $\times$ (e.g., WP4, WP6 in Figure~\ref{fig:heatmap-elpnorm}) indicates that certain starting points consistently fail regardless of random seed, while others (e.g., WP9) succeed reliably. This pattern suggests that starting point quality is a stronger determinant of success.

Table~\ref{tab:variance} confirms this quantitatively. Across 9 randomly chosen theorems with non-trivial variance, the average inter-WP std (3.36) exceeds the average intra-WP std (2.41), indicating that the variation due to different starting points is larger than the variation from repeated runs of the same starting point. This supports the conclusion that selecting a good wrong proof as the refinement starting point matters more than simply rerunning the same starting point multiple times.

\begin{table}[t]
    \centering
    \resizebox{0.3\textwidth}{!}{%
    \begin{tabular}{lcc}
        \toprule
        & Intra-WP std & Inter-WP std \\
        \midrule
        T1 & 1.86 & 5.59 \\
        T2 & 0.80 & 2.58 \\
        T3 & 7.02 & 8.66 \\
        T4 & 0.00 & 0.00 \\
        T5 & 0.08 & 0.49 \\
        T6 & 5.42 & 6.54 \\
        T7 & 1.22 & 0.49 \\
        T8 & 4.06 & 4.85 \\
        T9 & 1.24 & 1.00 \\
        \midrule
        Avg & 2.41 & 3.36 \\
        \bottomrule
    \end{tabular}
    }
    \caption{Intra-WP std vs.\ inter-WP std for theorems in the mathlib-valid.}
    \label{tab:variance}
\end{table}

\begin{table*}
  \centering
  \setlength{\tabcolsep}{4pt}
  \resizebox{0.8\textwidth}{!}{%
  \begin{tabular}{@{} l *{7}{c} @{\hskip 12pt} c @{}}
    \toprule
    \multirow{2}{*}{\textbf{Method}}
      & \multicolumn{7}{c}{\textbf{Pass rate (\%) per project}}
      & \multirow{2}{*}{\textbf{Avg.}} \\
    \cmidrule(lr){2-8}
      & Carleson & ConNF & FLT & Foundation & HepLean & Mathlib & Seymour & \\
    \midrule
    GPT-5-mini (pass@32)
      & 40.00 & 34.00 & 38.24 & 56.00 & 56.00 & 44.00 & 80.00 & 49.75 \\[2pt]
    DeepSeek-Prover-v2 (pass@32)
      & 46.00 & 34.00 & 50.00 & 56.00 & 50.00 & 48.00 & 78.00 & 51.71 \\[2pt]
    \midrule
    \textbf{Ours w/o AutoSolve}
      & \textbf{52.00} & \textbf{46.00} & \textbf{52.94}
      & \textbf{66.00} & \textbf{64.00} & \textbf{54.00}
      & \textbf{90.00} & \textbf{60.71} \\
    \quad {\footnotesize\textcolor{gray}{avg calls}}
      & {\footnotesize\textcolor{gray}{23.50}}
      & {\footnotesize\textcolor{gray}{27.92}}
      & {\footnotesize\textcolor{gray}{24.24}}
      & {\footnotesize\textcolor{gray}{17.94}}
      & {\footnotesize\textcolor{gray}{18.56}}
      & {\footnotesize\textcolor{gray}{24.20}}
      & {\footnotesize\textcolor{gray}{8.58}}
      & {\footnotesize\textcolor{gray}{20.71}} \\
    \bottomrule
  \end{tabular}%
  }
  \caption{Pass rate (\%) on \texttt{miniCTX-v2} across seven projects without \textsc{AutoSolve}. 
           The two baselines use pass@32. Our method additionally reports average LLM calls per theorem (\textcolor{gray}{gray}).}
  \label{tab:no-autosolve}
\end{table*}

\begin{figure}[t]
  \centering
  \includegraphics[width=0.48\textwidth]{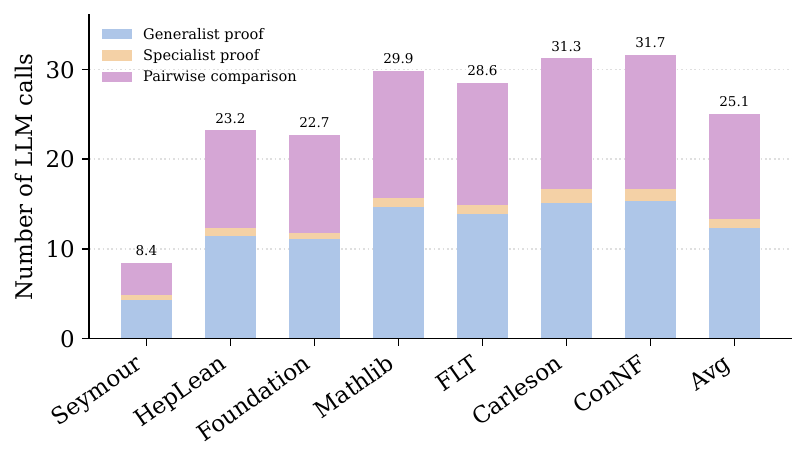}
  \caption{LLM calls breakdown under 32 average LLM calls on miniCTX-v2 test.}
  \label{fig:call_break}
\end{figure}

\begin{figure}[t]
  \centering
  \includegraphics[width=0.46\textwidth]{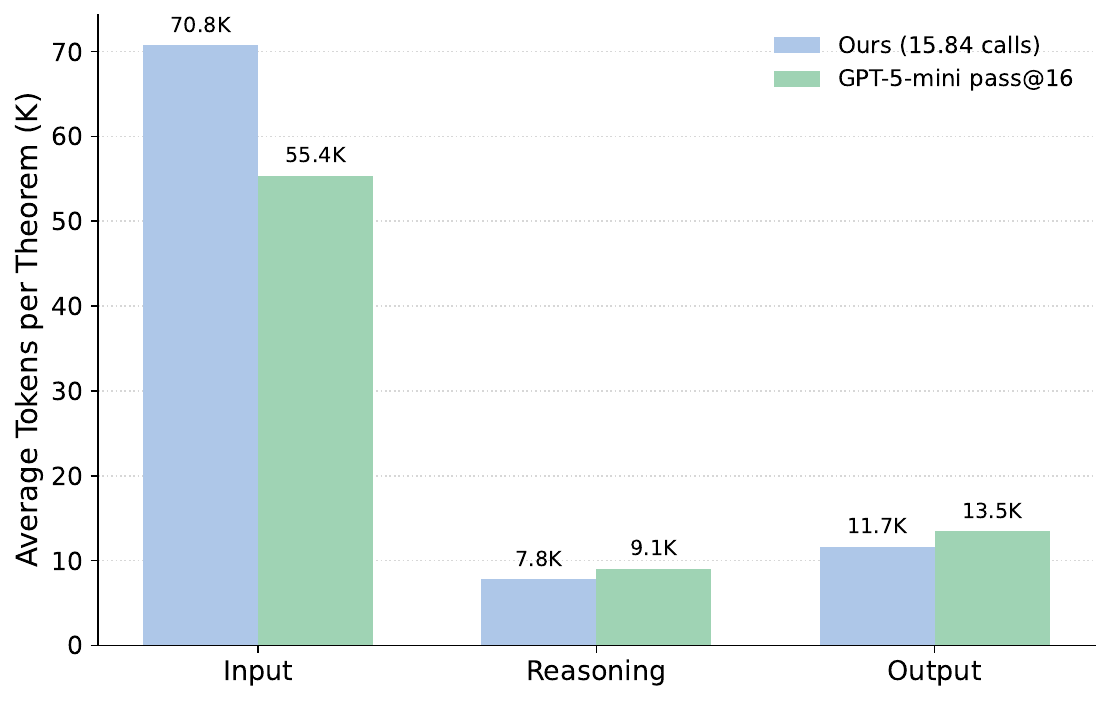}
  \caption{Average token consumption per theorem on HepLean under nearly matched LLM call budgets.}
  \label{fig:token_comparison}
\end{figure}

\subsection{LLM Calls Breakdown and Token Comparison}
\label{sec:call_breakdown}
Figure~\ref{fig:call_break} shows the average number of LLM calls per theorem. Pairwise comparison calls account for the largest part across all projects, but their cost is low as the judge produces only a short binary decision. The total varies from 8.4 calls on Seymour, where many theorems are resolved early, to over 31 on Carleson and ConNF, showing greater need for refinement on harder projects.

Figure~\ref{fig:token_comparison} shows the average token usage per theorem on HepLean project under similar number of LLM calls. Our method uses more input tokens because it includes compiler feedback, but fewer reasoning and output tokens because comparison call outputs short answers. At the same time, our method achieves better performance than the baseline (64\% vs.\ 56\%).

\begin{figure}[t]
  \centering
  \includegraphics[width=0.47\textwidth]{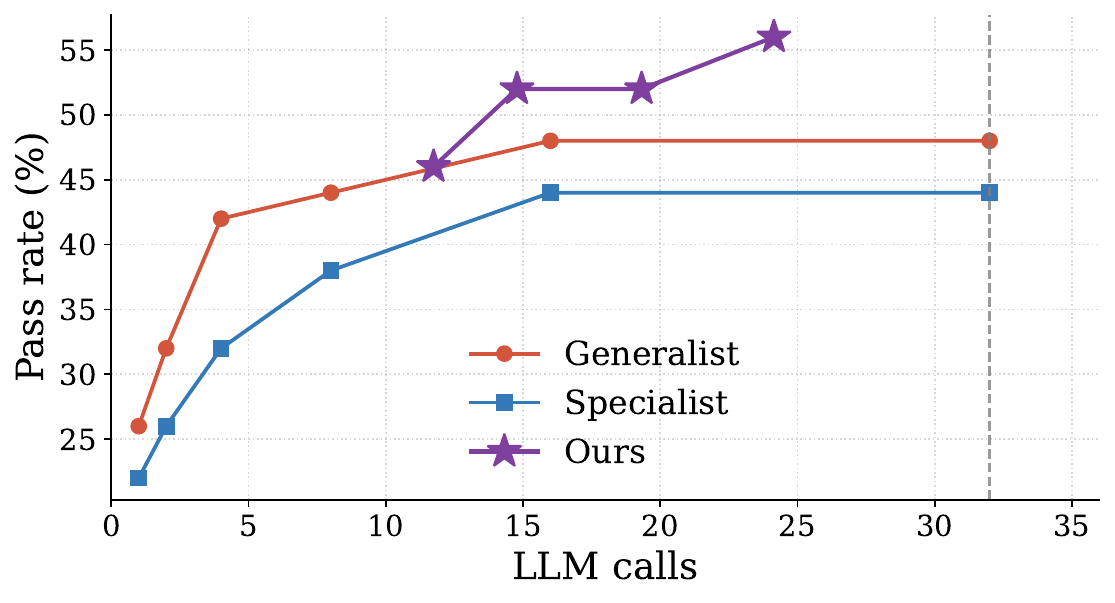}
  \caption{Pass rate (\%) \emph{vs.} LLM calls on HepLean with \textbf{GPT-5.4-mini} as the Generalist and \textbf{Goedel-Prover-V2-8B} as the Specialist.}
  \label{fig:goedel54}
\end{figure}

\subsection{Additional Model Pair: GPT-5.4-mini and Goedel-Prover-V2-8B}
\label{app:goedel}
We further evaluate robustness by replacing both the generalist and the specialist, using GPT-5.4-mini as the generalist and Goedel-Prover-V2-8B as the specialist. Figure~\ref{fig:goedel54} plots pass rate against LLM calls on HepLean. Both individual models plateau quickly, the generalist at 48\% and the specialist at 44\%, while our method continues to improve with additional calls, reaching over 56\%. This confirms that the complementary exploration strategy benefits from model diversity in general, rather than relying on the specific strengths of any single model pair.

\subsection{Results Without AutoSolve}
\label{sec:no-autosolve}

To separate the effect of \textsc{AutoSolve} from our adaptive search, we run our method with \textsc{AutoSolve} disabled. Table~\ref{tab:no-autosolve} compares this setting with the pass@32 baselines. Without \textsc{AutoSolve}, our method reaches an average pass rate of 60.71\%, compared with 49.75\% for GPT-5-mini and 51.71\% for DeepSeek-Prover-v2. It also achieves the highest pass rate on all seven projects while using 20.71 LLM calls per theorem on average. These results show that the main performance gain and the lower call usage are not from \textsc{AutoSolve}. Proofs found by \textsc{AutoSolve} can be solved by LLMs within very few calls.

\subsection{Resampling Example}
\label{sec:resampling-example}

We give an example to show the indirect role of resampling in the search process. On \texttt{TileStructure.Forest.exists\_p\_of\_mem\_$\sigma$} from Carleson project, four refinement steps failed to improve the current proof, and resampling was triggered at iteration 5. The new proof was selected but was still not valid. After one more refinement step, the theorem was solved. This success is counted as refinement rather than resampling in
Table~\ref{tab:success_source}. This example shows that resampling can help the search leave a stuck state and give later refinement a better starting point.

\subsection{Analysis of Explicit Use of Source Context}
\label{sec:explicit-context}

We conduct an additional experiment to understand how our method uses information from the source context. For each theorem, we extract the names of declarations in the provided source context before the target theorem. We then use lexical matching to check whether the verified proof explicitly refers to any of these declarations. We only count declaration names that appear directly in the proof. Thus, this analysis does not capture implicit use through tactics such as \texttt{simp} or \texttt{aesop}.

We report two measures. The \emph{reference rate} is the percentage of verified proofs that use at least one such declaration. The \emph{average number of references} is the average number of different source context declarations used each proof. Table~\ref{tab:explicit-context} shows that 42.79\% of the proofs found by our method use at least one additional declaration from the source context. The rate is 40.30\% for the ground-truth proofs. Refinement has the highest reference rate at 62.30\% and also the highest average number of references. AutoSolve has a much lower reference rate, which is expected because it mainly handles simpler proofs with built-in tactics. Initial proofs from the Generalist also make frequent use of the source context, with a reference rate of 60.66\%. In comparison, initial proofs from the Specialist use fewer explicit references. One possible reason is that the Generalist is better at reasoning over the provided context, while the Specialist focuses more on Lean proof generation.

\begin{table}[t]
  \centering
  \setlength{\tabcolsep}{7pt}
  \resizebox{0.9\columnwidth}{!}{%
  \begin{tabular}{lcc}
    \toprule
    \textbf{Proof source}
      & \textbf{Ref. rate (\%)}
      & \textbf{Avg. \# refs} \\
    \midrule
    AutoSolve
      & 3.51 & 0.04 \\
    Initial Specialist
      & 42.11 & 0.53 \\
    Initial Generalist
      & 60.66 & 0.93 \\
    Refinement
      & \textbf{62.30} & \textbf{1.11} \\
    Resampling
      & 33.33 & 0.33 \\
    \midrule
    \textbf{Full method}
      & 42.79 & 0.69 \\
    Ground-truth proof
      & 40.30 & 0.57 \\
    \bottomrule
  \end{tabular}%
  }
  \caption{
    Explicit use of declarations from the source context.
    We count a reference when a verified proof names a declaration that appears
    in the preceding source context but not in the target theorem statement.
    Ref. rate is the percentage of proofs with at least one such reference.
    Avg. \# refs is the average number of distinct references per proof.
    Results are pooled over theorems from all seven projects.
  }
  \label{tab:explicit-context}
\end{table}

\subsection{Analysis of Judge Model Preference}
\label{app:judge}

Table~\ref{tab:model_winrate} reports the model selection rate across all pairwise comparisons in which both GPT-5-mini and DeepSeek-Prover-V2 successfully produce a candidate proof during initial generation or stagnation-triggered resampling. These comparisons are counted over all theorems, including unsolved ones. Across 391 total comparisons, GPT-5-mini is selected 58.6\% of the time, indicating a moderate but consistent preference. This advantage is most pronounced on FLT (78.1\%) and HepLean (71.1\%), where GPT-5-mini's candidates are substantially favored. On other projects such as Carleson, Seymour, and Foundation, the selection rates are closer, suggesting that the relative strength of the two models is task dependent. These results indicate that employing a multi-model generation strategy is beneficial: although GPT-5-mini is preferred overall, DeepSeek-Prover-V2 is still
selected in over 40\% of comparisons, confirming that it provides complementary proof candidates. 

\begin{table}[t]
\centering
\resizebox{0.95\columnwidth}{!}{%
\begin{tabular}{lccc}
\toprule
\textbf{Project} & \textbf{\#Comparisons} & \textbf{GPT-5-mini (\%)} & \textbf{DeepSeek-Prover-V2 (\%)} \\
\midrule
Carleson    &  49 & 53.1 & 46.9 \\
ConNF       &  73 & 54.8 & 45.2 \\
FLT         &  32 & 78.1 & 21.9 \\
Foundation  &  47 & 55.3 & 44.7 \\
HepLean     &  45 & 71.1 & 28.9 \\
Mathlib     & 122 & 55.7 & 44.3 \\
Seymour     &  23 & 52.2 & 47.8 \\
\midrule
Total       & 391 & 58.6 & 41.4 \\
\bottomrule
\end{tabular}%
}
\caption{Model selection rate in pairwise comparisons, pooled over all
theorems. Each comparison occurs when both GPT-5-mini and
DeepSeek-Prover-V2 produce a candidate proof and the judge selects the
better starting point.}
\label{tab:model_winrate}
\end{table}

\begin{table*}[t]
  \centering
  \footnotesize
  \setlength{\tabcolsep}{4pt}
  \begin{tabular}{@{} l *{7}{c} @{\hskip 10pt} c @{}}
    \toprule
    \multirow{2}{*}{\textbf{Component}}
      & \multicolumn{7}{c}{\textbf{Number of theorems proved per project}}
      & \multirow{2}{*}{\textbf{Avg.}} \\
    \cmidrule(lr){2-8}
      & Carleson & ConNF & FLT & Foundation & HepLean & Mathlib & Seymour & \\
    \midrule
    AutoSolve
      & 0 & 7 & 11 & 20 & 4 & 10 & 5 & 8.14 \\
    Init (Generalist)
      & 16 & 2 & 0 & 5 & 10 & 3 & 25 & 8.71 \\
    Init (Specialist)
      & 4 & 1 & 3 & 1 & 4 & 3 & 3 & 2.71 \\
    Resample (Generalist)
      & 0 & 0 & 0 & 0 & 1 & 0 & 0 & 0.14 \\
    Resample (Specialist)
      & 0 & 0 & 0 & 0 & 1 & 0 & 1 & 0.29 \\
    Refine (Generalist)
      & 5 & 5 & 4 & 3 & 7 & 8 & 6 & 5.43 \\
    Refine (Specialist)
      & 2 & 7 & 1 & 1 & 5 & 3 & 4 & 3.29 \\
    \midrule
    \textbf{Total}
      & \textbf{27} & \textbf{22} & \textbf{19} & \textbf{30} & \textbf{32} & \textbf{27} & \textbf{44} & \textbf{28.71} \\
    \bottomrule
  \end{tabular}
  \caption{Detailed breakdown of theorems proved by each component on \texttt{miniCTX-v2 test}.}
  \label{tab:detailed-breakdown}
\end{table*}

\section{Implementation Details}
\label{app:impl}
 
We use \texttt{lean-interact}~\citep{leaninteract} to interact with the Lean~4 REPL.
For each theorem, the surrounding source context is validated by running it through the REPL (timeout = 360), producing a cached environment $\mathcal{E}$ against which all proof attempts are verified (timeout = 120). Unless otherwise specified, all main results are reported from a single run for each method and budget. 

\paragraph{Context Use.} Following miniCTX~\citep{hu2024minictx} and RLMEval~\citep{poiroux-etal-2025-rlmeval}, $C$ is the source content before the target theorem in the project file. It includes imports, local definitions, project-specific lemmas, namespaces, and notation. It comes directly from the original Lean repository. The generalist receives the full context because it can handle long inputs. The specialist has a shorter context window, so we keep the beginning and the end of $C$ and remove the middle part, following RLMEval~\citep{poiroux-etal-2025-rlmeval}. We set the context length to 8192 for the specialist, and increasing it further does not improve performance.

\paragraph{Lean version.} We use v4.16.0 for miniCTX-v2 and v4.7.0-rc2 for RLMEval-FLT3.
 
\paragraph{AutoSolve.} The \textsc{AutoSolve} module attempts to close the theorem before any LLM call using two classes of tactics:

(i)~deterministic closers (\texttt{rfl}, \texttt{simp}, \texttt{omega}, \texttt{linarith}, \texttt{norm\_num}, \texttt{ring}, \texttt{decide}, etc.) that close the goal directly, and

(ii)~search-based tactics (\texttt{exact?}, \texttt{hint}) that query Lean's library search, return ``Try this: \ldots'' suggestions, and verify each suggestion against the compiler.

\paragraph{Parsing Pairwise Decisions.} The pairwise judge is prompted to return \texttt{\{``choice'': ``A''\}} or \texttt{\{``choice'': ``B''\}}.
Responses are parsed with cascading fallbacks: JSON parsing, regex extraction, single-character detection, and if all fail, defaulting to the current proof (choice~A).
 
\paragraph{Error Extraction.} When a proof fails verification, compiler errors are matched to specific tactic lines by aligning error positions with proof source lines. Each error is formatted with one line of surrounding context and the error message, producing a compact line-level report used by both the repair model and the comparison judge.

\paragraph{Hardware.} We run the self-hosted specialist model on a server with two NVIDIA L40S GPUs, each with 48GB of memory. The generalist model is accessed through an API.
 

\section{Prompt Templates}
\label{app:prompts}

This section lists the prompt templates used for proof generation, refinement, and pairwise comparison.
Each prompt includes the theorem statement and the surrounding Lean project context.
Refinement and comparison prompts also include compiler feedback from failed proof attempts.

For RLMEval-FLT3, each theorem is paired with a natural language proof.
We include this natural language proof in the generation, refinement, and pairwise comparison prompts, while keeping the rest of the search procedure unchanged.

\subsection{Specialist Proof Generation}
\begin{tcolorbox}[colback=gray!5, colframe=gray!50, fontupper=\small\ttfamily, breakable]
Complete the following Lean 4 code with explanatory comments preceding each line of code:\\[4pt]
\textasciigrave\textasciigrave\textasciigrave lean4\\
\{context\}\\
\{declaration\}\\
\end{tcolorbox}

\subsection{Generalist Proof Generation}
\begin{tcolorbox}[colback=gray!5, colframe=gray!50, fontupper=\small\ttfamily, breakable]
Here is the Lean 4 context:\\
\textasciigrave\textasciigrave\textasciigrave lean4\\
\{context\}\\
\textasciigrave\textasciigrave\textasciigrave\\[4pt]
Using this context, prove the following theorem in Lean 4.\\
Theorem statement:\\
\{declaration\}\\
Start your response like this:\\
\textasciigrave\textasciigrave\textasciigrave lean4\\
:= by
\end{tcolorbox}

\subsection{Error-Guided Repair}
\begin{tcolorbox}[colback=gray!5, colframe=gray!50, fontupper=\small\ttfamily, breakable]
Here is the Lean 4 context:\\
\textasciigrave\textasciigrave\textasciigrave lean4\\
\{context\}\\
\textasciigrave\textasciigrave\textasciigrave\\[4pt]
Here is the target theorem statement:\\
\{declaration\}\\[4pt]
Here is an attempted Lean 4 proof body but contains errors:\\
\textasciigrave\textasciigrave\textasciigrave lean4\\
\{wrong\_proof\}\\
\textasciigrave\textasciigrave\textasciigrave\\[4pt]
The proof has the following errors:\\
\{errors\}\\[4pt]
Based on the context and errors, provide a new corrected Lean 4 proof.\\
Please output a new lean4 proof ONLY, no other explanations.
\end{tcolorbox}

\subsection{Pairwise Proof Comparison}
\begin{tcolorbox}[colback=gray!5, colframe=gray!50, fontupper=\small\ttfamily, breakable]
You are an expert in Lean 4. Your task is to compare two proofs that both failed verification, select the one that has better quality and is more suitable as the starting point for refinement.\\[4pt]
Here is the Lean 4 context:\\
\textasciigrave\textasciigrave\textasciigrave lean4\\
\{context\}\\
\textasciigrave\textasciigrave\textasciigrave\\[4pt]
Here is the theorem declaration:\\
\textasciigrave\textasciigrave\textasciigrave lean4\\
\{declaration\}\\
\textasciigrave\textasciigrave\textasciigrave\\[4pt]
--- Proof A ---\\
\textasciigrave\textasciigrave\textasciigrave lean4\\
\{proof\_A\}\\
\textasciigrave\textasciigrave\textasciigrave\\
Errors for Proof A:\\
\{errors\_A\}\\[4pt]
--- Proof B ---\\
\textasciigrave\textasciigrave\textasciigrave lean4\\
\{proof\_B\}\\
\textasciigrave\textasciigrave\textasciigrave\\
Errors for Proof B:\\
\{errors\_B\}\\[4pt]
Compare the two proofs above. Consider the following criteria:\\
Fixability: Which error message is easier to resolve?\\
Progress: Which proof leaves fewer or simpler remaining subgoals?\\
Strategy: Which proof uses a sounder overall approach?\\[4pt]
Respond with ONLY this JSON: \{"choice": "A"\} or \{"choice": "B"\}, no other text.
\end{tcolorbox}

\section{Effect of Context Length on Success Rate}
\label{app:context_length}

\begin{figure*}[p]
    \centering
    \setlength{\tabcolsep}{2pt}
    \renewcommand{\arraystretch}{0.9}

    \begin{tabular}{cc}
        \includegraphics[width=0.5\textwidth]{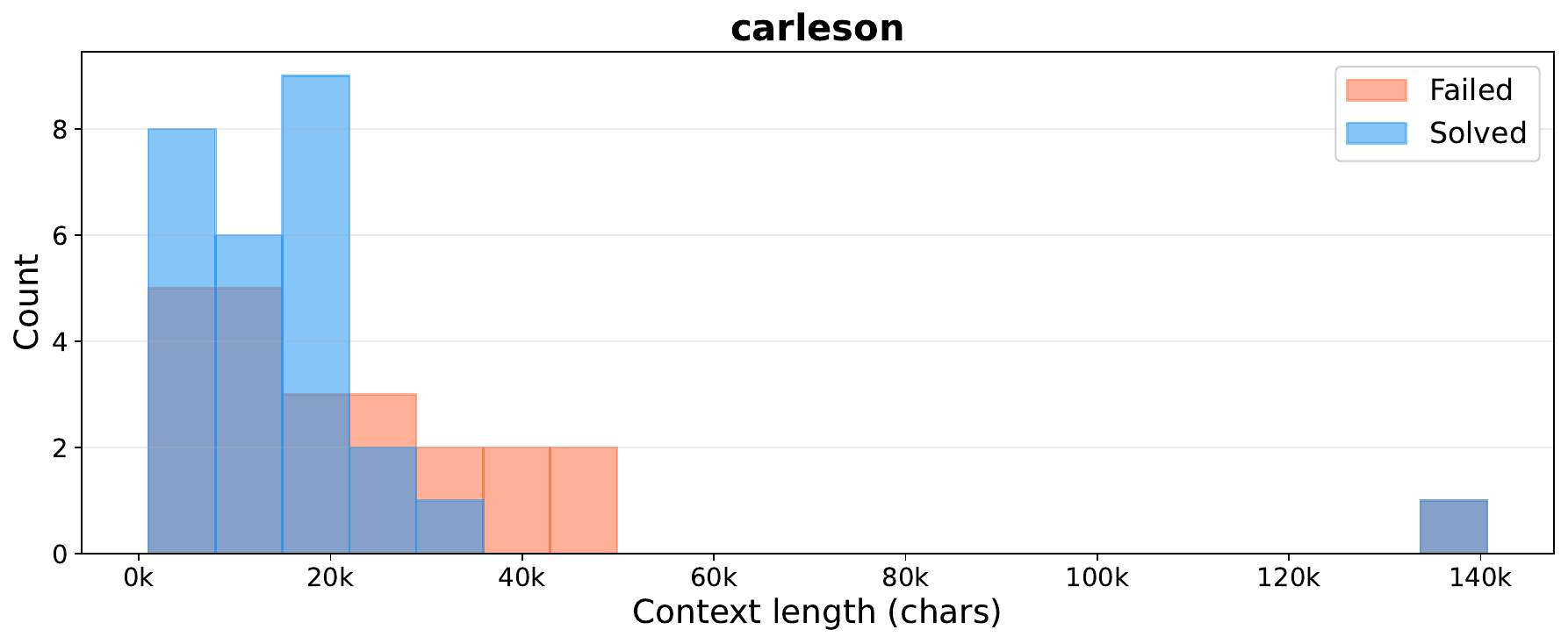} &
        \includegraphics[width=0.5\textwidth]{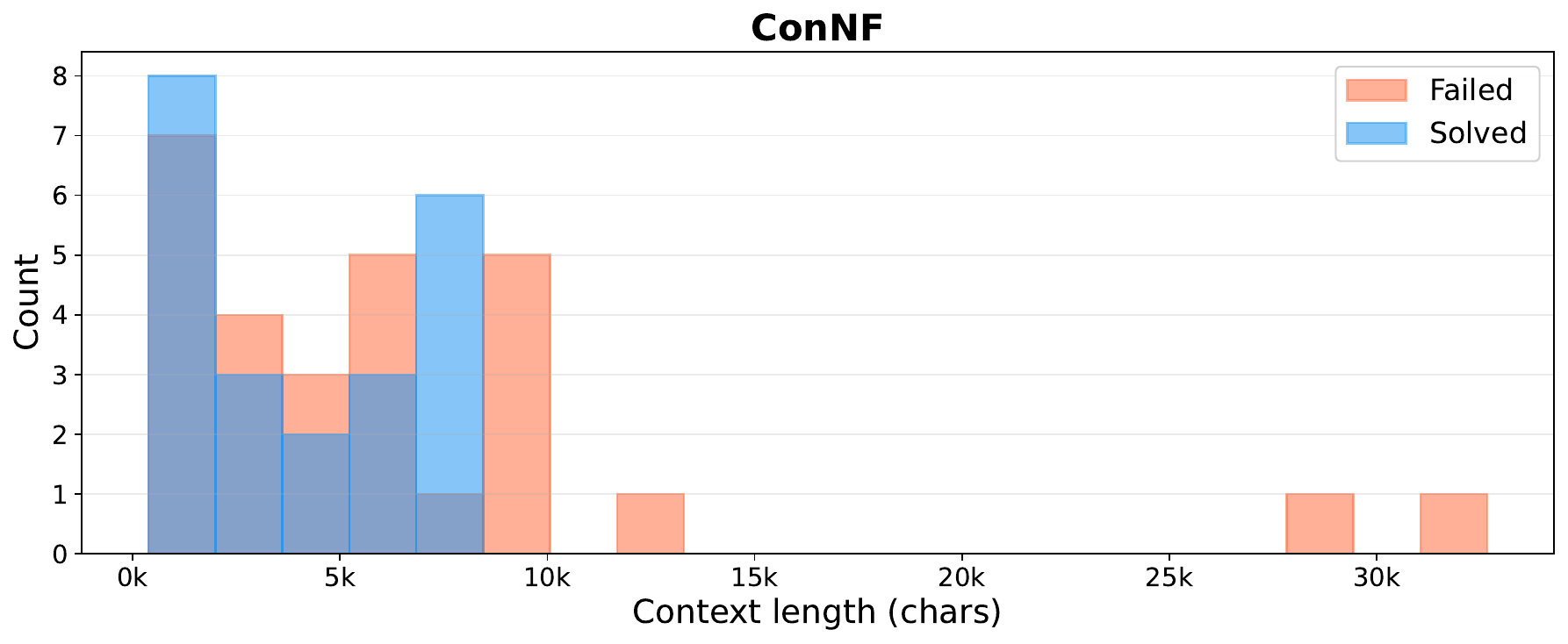} \\[-0.5em]

        \includegraphics[width=0.5\textwidth]{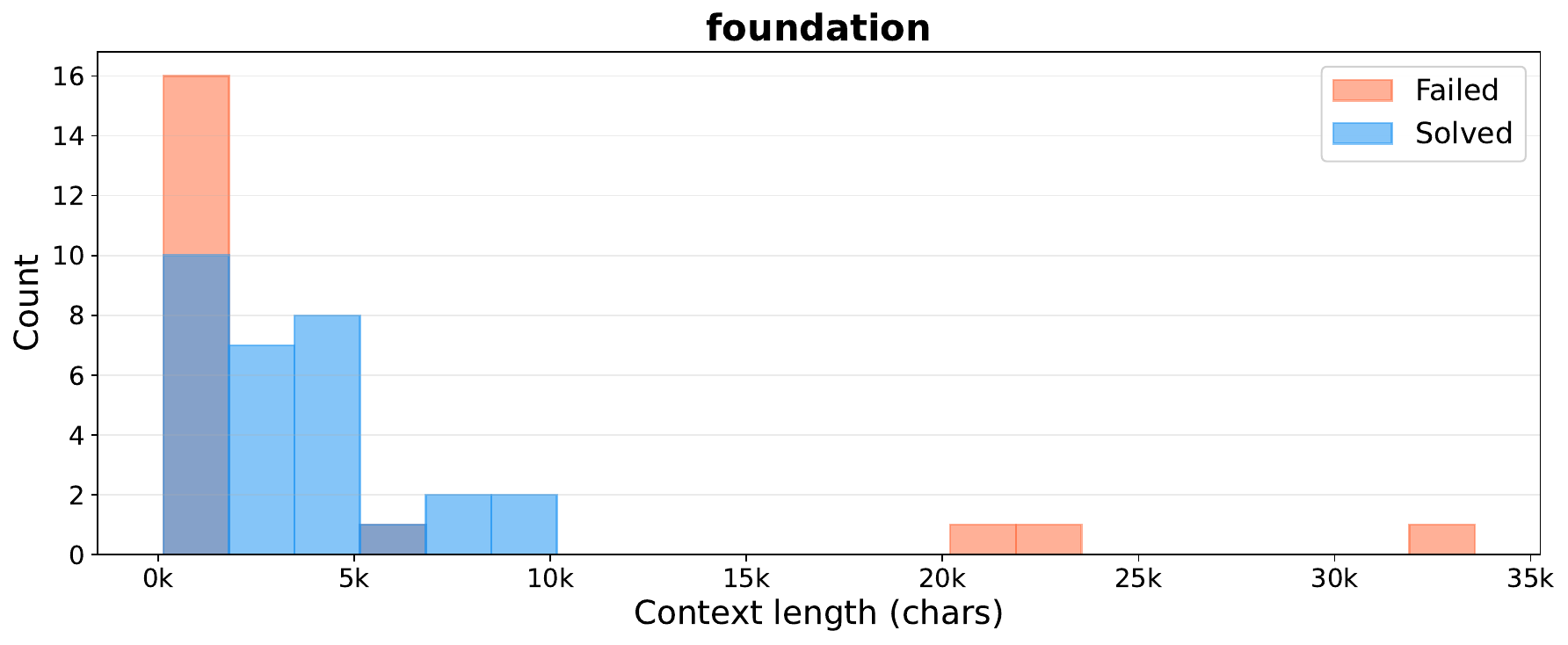} &
        \includegraphics[width=0.5\textwidth]{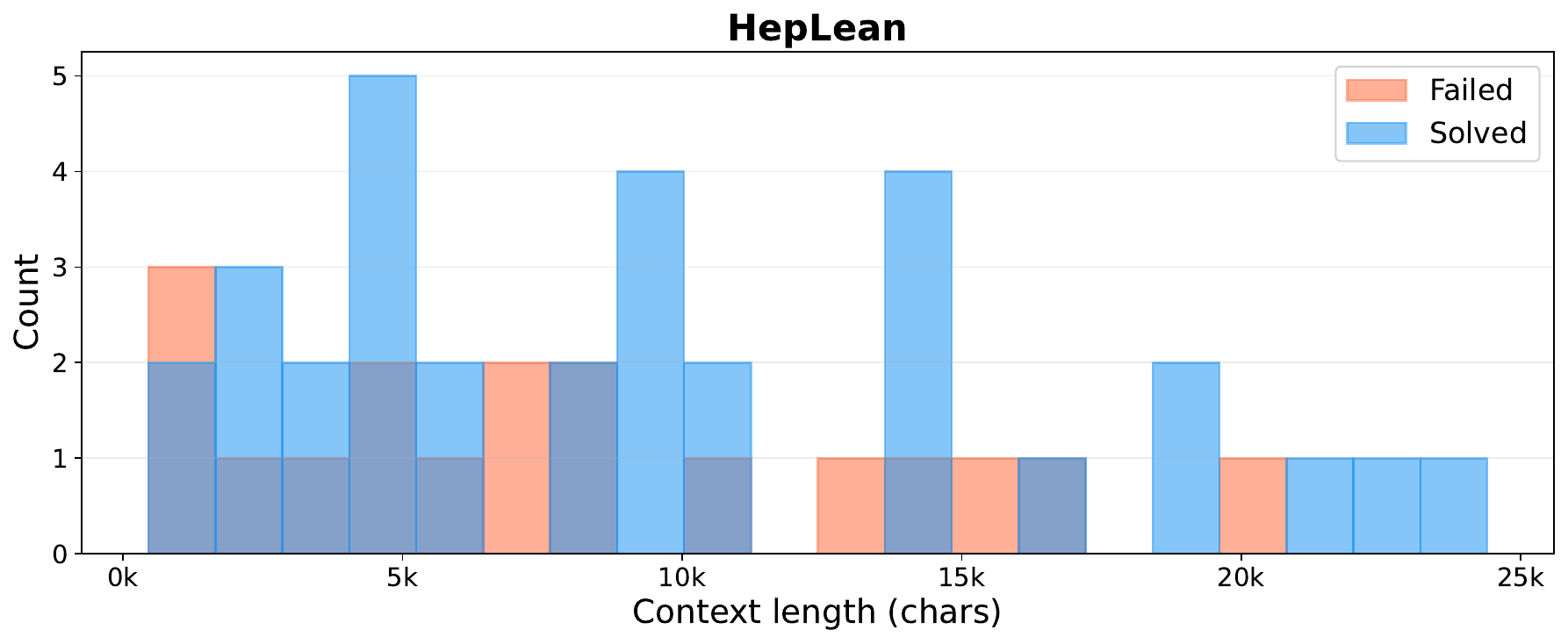} \\[-0.5em]

        \includegraphics[width=0.5\textwidth]{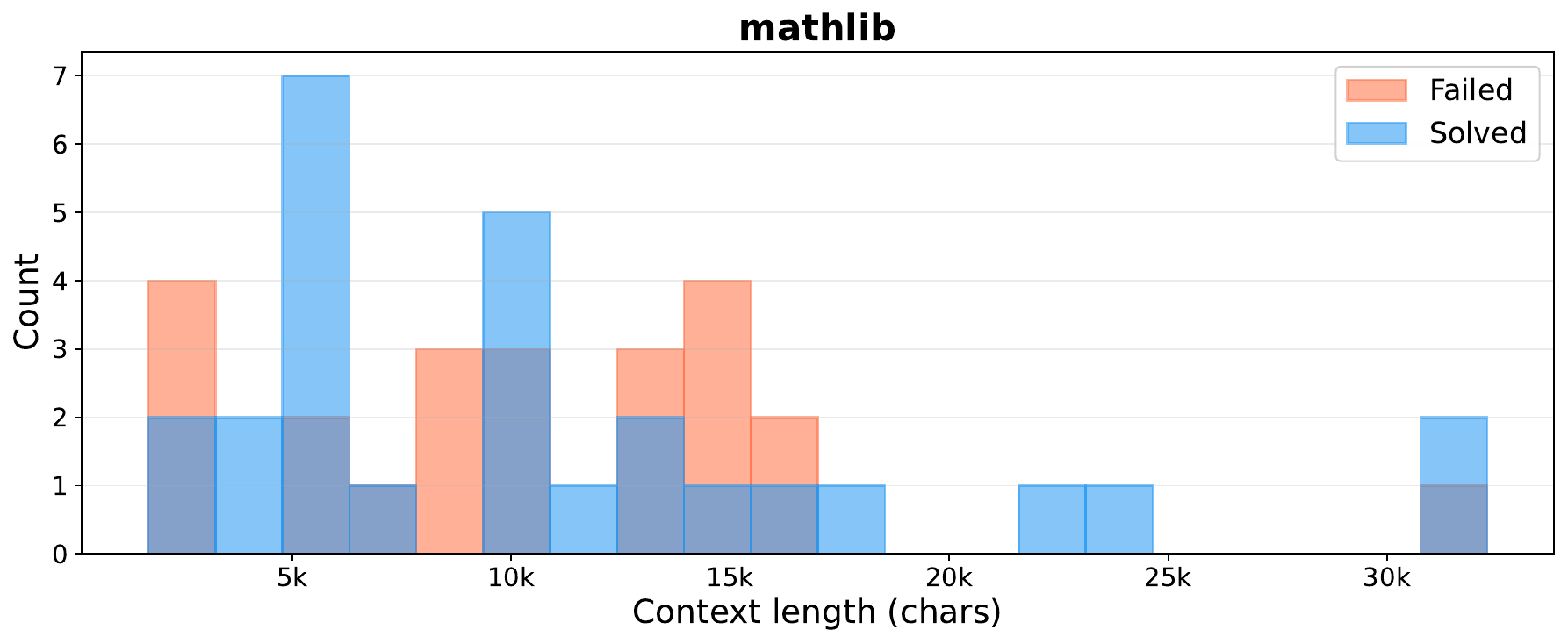} &
        \includegraphics[width=0.5\textwidth]{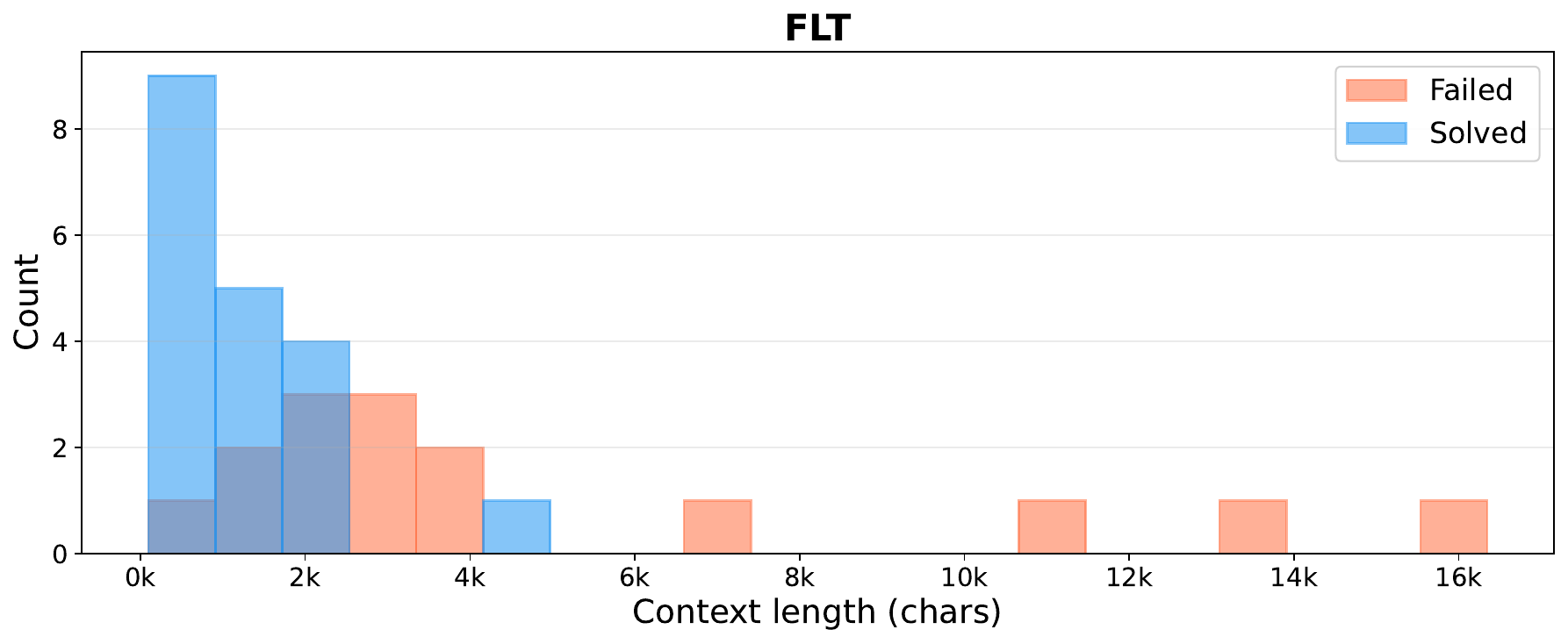} \\[-0.5em]

        \multicolumn{2}{c}{
        \includegraphics[width=0.5\textwidth]{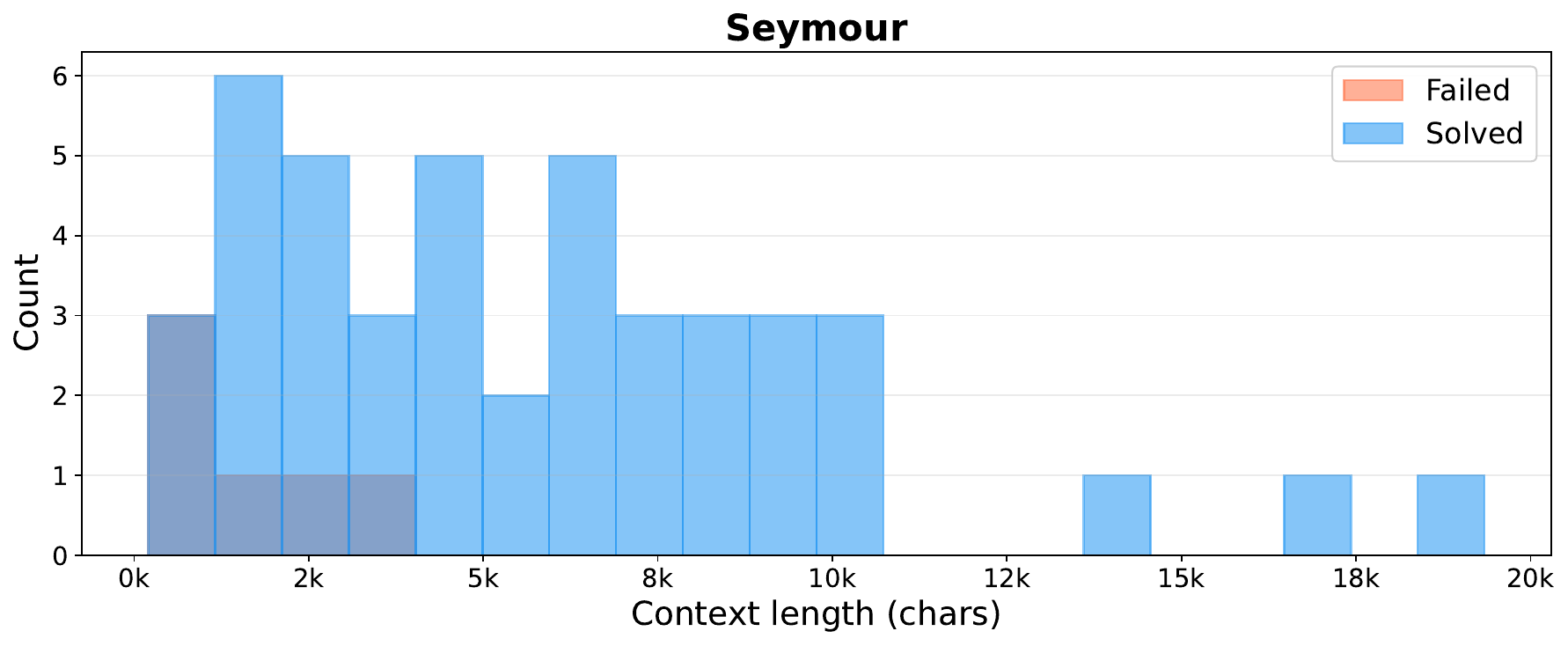}
        }
    \end{tabular}

    \caption{Distribution of proof success and failure by source context length across seven miniCTX-v2 projects.}
    \label{fig:ctx_histograms}
\end{figure*}

To understand how the surrounding Lean context influences proving difficulty, we analyze the relationship between source context length and proof success rate across all seven projects with max refinement $K = 16$. Figure~\ref{fig:ctx_histograms} shows the distribution of solved and failed theorems as a function of context length (in characters) for each project.

The seven projects have different context length distributions. FLT and Seymour have the shortest contexts, with almost all theorems below 20k characters. Carleson is the main outlier, with some contexts longer than 140k characters. In most projects, longer contexts are more often linked to failed proofs. This pattern is strongest in Carleson, where most theorems above 40k characters are not solved. A similar trend appears in Foundation and ConNF. In FLT and Seymour, the results are more balanced because the contexts are generally short. Overall, longer source contexts are often associated with harder proofs. They may include more irrelevant information and also tend to appear in more complex theorems.

\section{Artifact Use}
\label{app:artifact}

\paragraph{Licenses and terms of use.}
We use publicly available benchmarks and tools, including miniCTX-v2, RLMEval-FLT3, Lean 4, and lean-interact, under their released terms of use and licenses. We use the associated Lean projects only for research evaluation and do not redistribute modified project artifacts. For API-based generalist models and self-hosted specialist models, we follow the corresponding provider terms and model licenses.

\paragraph{Intended use.}
Our use of these artifacts is consistent with their intended research and evaluation use. We use theorem proving benchmarks to evaluate proof search systems, Lean projects as proof environments, and models for proof generation, refinement, and comparison.

\paragraph{Privacy and offensive content.}
The artifacts used in this work are theorem proving benchmarks, Lean source files, compiler feedback, and informal mathematical proofs. They are not datasets about individual people and are not expected to contain personally identifying information or offensive content.

\end{document}